\documentclass{article}

\PassOptionsToPackage{numbers,compress}{natbib}
\usepackage[main,final]{neurips_2026}

\usepackage[utf8]{inputenc}
\usepackage[T1]{fontenc}
\usepackage{hyperref}
\usepackage{url}
\usepackage{booktabs}
\usepackage{amsfonts}
\usepackage{amsmath}
\usepackage{amssymb}
\usepackage{microtype}
\usepackage{xcolor}
\usepackage{graphicx}
\usepackage{tabularx}
\usepackage{siunitx}
\usepackage{pgfplots}
\pgfplotsset{compat=1.18}
\usepackage{enumitem}
\usepackage{algorithm}
\usepackage{algpseudocode}
\usepackage{wrapfig}

\usepackage{longtable}

\newcommand{\method}{VHG}

\let\originalbibitem\bibitem
\renewcommand{\bibitem}[2][]{%
  \color{black}%
  \originalbibitem[#1]{#2}}

\title{Verifier-Backed Hard Problem Generation for Mathematical Reasoning}

\author{Yuhang Lai\textsuperscript{*\,1,2} \quad
Jiazhan Feng\textsuperscript{*$\dagger$\,3} \quad
Yee Whye Teh\textsuperscript{4} \quad
Ning Miao\textsuperscript{$\ddagger$\,1,2}\\[1mm]\textsuperscript{1}Department of Data Science, City University of Hong Kong\\\textsuperscript{2}Hong Kong Institute of AI for Science, City University of Hong Kong\\\textsuperscript{3}School of Intelligence Science and Technology, Peking University\\\textsuperscript{4}Department of Statistics, University of Oxford
}

\begin{document}

\maketitle
\begingroup

\renewcommand{\thefootnote}{*}\footnotetext{ Equal contribution.}
\renewcommand{\thefootnote}{\ensuremath{\dagger}}\footnotetext{ Work done while Jiazhan Feng was visiting the Department of Statistics, University of Oxford.}
\renewcommand{\thefootnote}{\ensuremath{\ddagger}}\footnotetext{ Correspondence to: Ning Miao, \href{mailto:ningmiao@cityu.edu.hk}{ningmiao@cityu.edu.hk}.}
\endgroup
\begin{abstract}
Large Language Models (LLMs) demonstrate strong capability in solving scientific and mathematical problems, yet they struggle to produce valid and challenging novel problems---an essential component for advancing LLM training and enabling autonomous scientific research. 
Existing problem generation approaches either depend on expensive human expert involvement or adopt naive self-play paradigms, which frequently yield invalid problems due to reward hacking.
This work introduces VHG, a verifier-backed hard problem generation framework built upon three-party self-play. By integrating an independent verifier into the conventional setter-solver duality, our design constrains the setter's reward to be jointly determined by problem validity (evaluated by the verifier) and difficulty (assessed by the solver). We instantiate two verifier variants: a Hard symbolic verifier and a Soft LLM-based verifier, with evaluations conducted on indefinite integral tasks and general mathematical reasoning tasks. Experimental results show that VHG substantially outperforms all baseline methods by a clear margin. Our code is available at \url{https://github.com/Miaow-Lab/VHG}.
\end{abstract}

\section{Introduction}
\label{sec:intro}
Large language models~(LLMs) have achieved expert-level ability in solving scientific problems.
For example, OpenAI's o1 surpassed PhD-expert baselines on GPQA-Diamond \citep{rein2024gpqa,openai2024learning}, while AlphaGeometry \citep{trinh2024alphageometry} and AlphaProof \citep{hubert2026alphaproof} demonstrated olympiad-level mathematical reasoning, making them very useful in doing research under human supervision.
In scientific research, raising a meaningful new question is at least as critical as solving an existing one.
This raises a natural question: can LLMs truly generate valid new questions, enabling them to complete the research cycle and achieve fully autonomous research?

Generating valid new questions is not only pivotal to scientific research but also to the training of LLMs themselves. 
For example, \citet{gao2025promptcurriculum} found that data difficulty is one of the primary factors influencing the performance of LLMs after post-training. 
However, current post-training paradigms mostly rely on static human-written datasets (e.g., MATH \citep{hendrycks2021math}), or offline transformations and post-training recipes (e.g., MetaMath \citep{yu2024metamath}, WizardMath \citep{luo2024wizardmath}, ToRA \citep{gou2024tora}, and DeepSeekMath \citep{shao2024deepseekmath}). 
To continuously increase the difficulty of training data, we need to hire domain experts either to compose new problems or modify existing ones, which is not only slow and costly, but also limits LLMs' ability to the human level.

Numerous previous efforts have been made on generating difficult problems. 
For example, PromptCoT \citep{zhao2025promptcot}, CHASE \citep{patel2025chase}, and MathSmith \citep{zhan2025mathsmith} use human-designed scaffolds, independently verifiable components, or predefined difficulty strategies to build new problems. 
MathFusion combines old problems through sequential, parallel, and conditional fusion strategies \citep{pei2025mathfusion}. 
Though effective in certain tasks, these methods are still upper-bounded by human design. 
Self-play based methods almost eliminate reliance on human expertise. 
They use a setter LLM to propose new problems, and another solver LLM to measure the difficulty of the proposed problems.
Then, the negative problem accuracy is used as a reward for the setter, so that it will learn to generate harder and harder problems.
Though looks elegant, vanilla self-play has a critical flaw, that is the proxy reward of problem difficulty can be easily hacked by generating invalid problems, where the solver has zero accuracy, which provides high rewards to the setter. 

\begin{wrapfigure}[17]{r}{0.55\textwidth}
\centering
\vspace{-10pt}
\includegraphics[width=0.53\textwidth]
{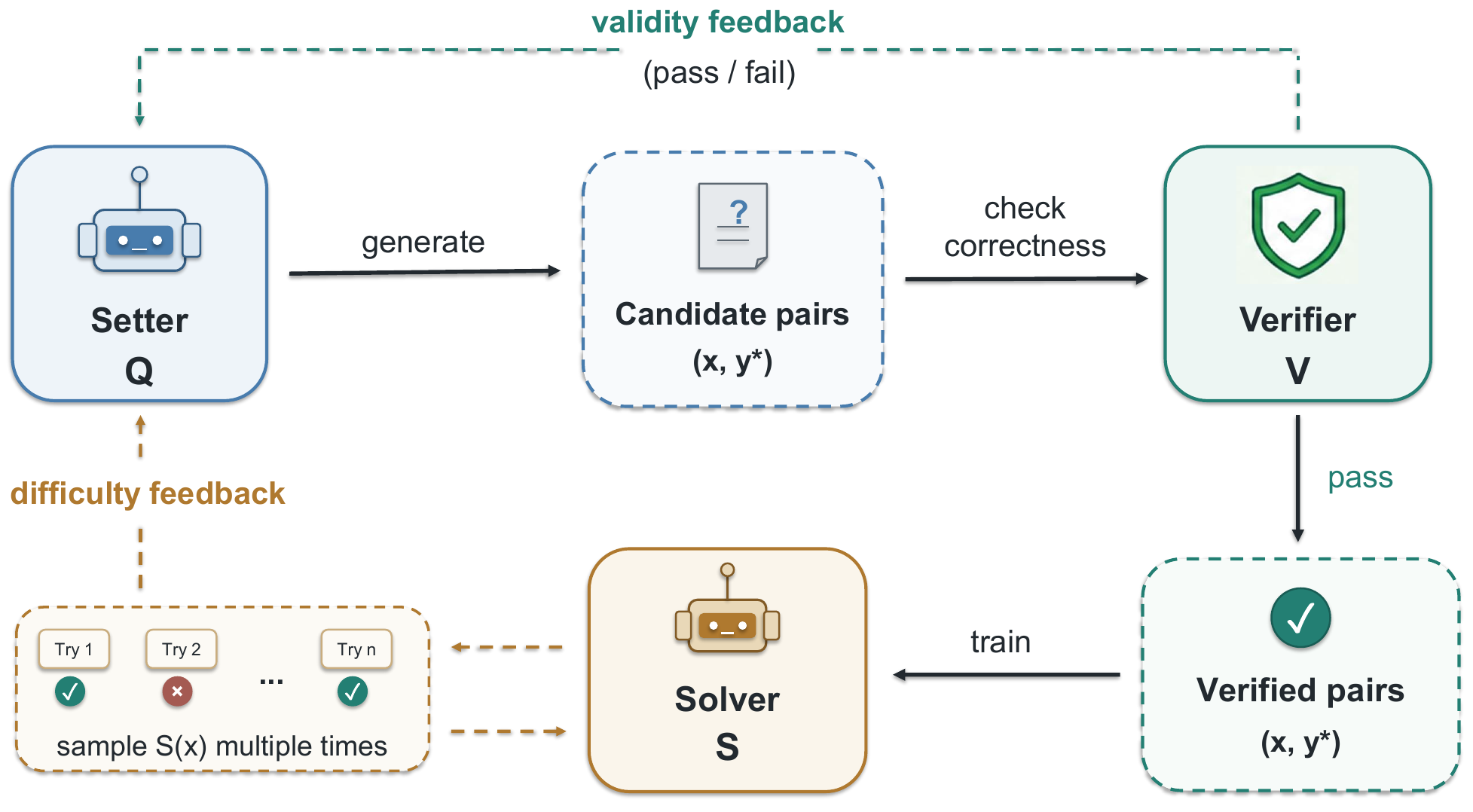}
\caption{{\method} framework. The setter proposes problem-reference pairs, the verifier gates validity, and accepted pairs are scored by solver difficulty for training and challenge construction.}
\label{fig:framework}
\end{wrapfigure}

In this paper, we address this problem by adding an independent verifier to the two-party game in self-play.
Specifically, we propose \emph{{\method}: verifier-backed hard problem generation}: a setter proposes both the problem and solution as a pair $(x,y^\star)$, and a verifier accepts or rejects the pair based on its correctness. 
This design achieves two key goals:
On the solver side, it eliminates the noisy training signals introduced by invalid problem-reference pairs.
More importantly, on the setter side, for a generated problem, a high reward (i.e., low solver accuracy) can only result from the problem's true difficulty, making it impossible for the setter to hack the reward system by generating invalid problems.

We explore two types of verifiers: Hard and Soft. 
Hard verifiers leverage symbolic verification mechanisms to provide nearly 100\% reliable verifications. 
Soft verifiers use LLMs to check the correctness of the step-by-step problem generation process. Though less accurate than hard verifiers, soft verifiers enable the framework to work on broader domains.
To validate the concept of {\method}, we first focus on the task of indefinite integral, a self-contained environment, allowing us to examine the inner working mechanisms of {\method}.
We also test {\method} with a soft verifier in the general math domain, where exact verification is impractical, demonstrating its generalization potential.

Our experiments show that {\method} produces problem-reference pairs that are both valid and challenging, proving useful for both solver training and challenge dataset construction.
After three iterations, {\method} improves pass@1 over the base model on AntiderivBench Qualifier/Competition and the Integral Stress Test by 22.0, 22.2, and 24.5 percentage points, respectively, outperforming baseline approaches including R-Zero, which is the current most typical framework. 
For general math, {\method} enhances performance across a wide range of benchmarks including MATH, AMC, Minerva, Olympiad, and AIME24-26: raising overall pass@1 accuracy from 56.8\% to 69.0\% and also outperforming baselines by a substantial margin.
Beyond performance gains, we find that even though both the setter and the solver are based on Qwen3-4B, {\method} can generate problems that challenge larger models (Qwen3-8B, 14B, and 32B), indicating that stronger models can benefit from training with data generated by weaker models.

In summary, the contributions of this paper are as follows:
\begin{itemize}[topsep=0pt,
  partopsep=0pt,
  itemsep=2pt,
  parsep=0pt,
  leftmargin=*]
    \item We design a novel three-party self-play framework for generating challenging problem, incorporating an additional verifier module. This framework avoids the generation of invalid problems, which would otherwise lead to a collapse of the training pipeline.
    \item {We empirically evaluate {\method} with both hard and soft verifiers, showing that it generates novel, challenging problems that in turn boost performance across multiple benchmarks.}
    \item We perform a comprehensive analysis of {\method}'s working mechanism, highlighting the critical role of the verifier in ensuring the quality of generated problems.
\end{itemize}

\section{Method: The {\method} Framework}
\label{sec:framework}
Verifier-backed hard problem generation ({\method}) is a three-party self-play framework designed to generate mathematical problems that are valid, novel, and challenging. It extends the standard setter--solver self-play paradigm by introducing a verifier: the setter proposes a problem-reference pair, the solver provides feedback on its empirical difficulty, and the verifier validates the pair's correctness. During training, generated pairs are first checked for validity, and only verifier-accepted pairs are scored for difficulty or used to train the solver. This ensures that invalid or underspecified pairs are excluded from downstream processes, mitigating reward hacking and ensuring the quality of generated problems.

\paragraph{Setter $\mathrm{Q}$.}
The setter $\mathrm{Q}$ is responsible for generating problem-reference pairs $(x,y^\star)$, where $x$ denotes a proposed mathematical problem and $y^\star$ is its reference solution. This extends the conventional self-play formulation, where the setter typically only needs to propose a problem without a reference solution. In our framework, the reference solution $y^\star$ is also part of the generated output. This inclusion enables verifiable validity checks and provides a target against which solver outputs can be evaluated. 
The setter's role is architecture-agnostic: it can modify existing problems, compose multiple seed problems, or synthesize entirely new problems from broader contextual information. 
In this work, we instantiate $\mathrm{Q}$ as an LLM conditioned on seed problem-reference pairs. 
Unlike fixed hand-designed problem generators, such as template libraries or rule-based perturbation systems, our LLM-based setter is trained with feedback from both the solver and the verifier. The training signal incentivizes the setter to generate problems that are not only difficult for the solver but also formally valid according to the verifier's criteria.

\paragraph{Solver $\mathrm{S}$.}
The solver $\mathrm{S}$ is another model whose failures reflect the difficulty of problems. Given a generated problem $x$, the solver produces solution attempts under a fixed sampling budget, and its empirical accuracy $\mathrm{Acc}_S(x, y^\star)$, defined as the proportion of correct solutions, serves as the difficulty signal in the {\method} framework. This role aligns with the solver's function in standard reinforcement-learning or self-play settings: it learns to solve problems, while its current limitations reveal which problems are sufficiently challenging. A key distinction in {\method} is that solver failure only contributes to the setter's training signal if the verifier first accepts the corresponding problem-reference pair. This decouples validity from difficulty, ensuring that the setter is not rewarded for generating invalid or underspecified problems that the solver cannot answer.

\paragraph{Verifier $\mathrm{V}$.}
The verifier $\mathrm{V}$ validates the generated pair $(x,y^\star)$ by checking whether it meets a predefined acceptance criterion. In this work, we instantiate two distinct types of verifiers--hard verifiers and soft verifiers to accommodate different task constraints and availability of formal validation tools. 
Hard verifiers are usually external computational checks to provide definitive validity judgments. For example, in the context of indefinite integrals, we use SymPy (a symbolic mathematics library) to verify that the proposed antiderivative correctly differentiates to the integrand. Hard verifiers offer high auditability and correctness guarantees when such formal checks are feasible.
In contrast, soft verifiers use LLMs to validate problem-reference pairs.
While soft verifiers may introduce noise compared with hard verifiers, they provide a generalizable solution that can be applied to a broader range of mathematical tasks where formal symbolic checks are unavailable or impractical.

\paragraph{Training.}
The introduction of the verifier addresses a critical limitation of vanilla self-play: without a validity check, the setter can exploit reward signals by generating invalid or underspecified problems that the solver fails to answer, leading to spurious difficulty. In {\method}, the setter's reward is conditioned on verifier acceptance, ensuring that solver failure only contributes to the reward if the problem-reference pair is valid. Formally, the setter's reward function is defined as:
\begin{equation}
R_{\mathrm{Q}}(x,y^\star)
=
\mathbf{1}_{[V(x,y^\star)=1]}\left(1-\mathrm{Acc}_S(x,y^\star)\right),
\label{eq:verify-score}
\end{equation}
where $\mathrm{Acc}_S(x,y^\star)$ is estimated under a fixed sampling protocol. If the verifier rejects the pair~(when $V(x,y^\star)=0$), the solver's failure does not contribute to the reward. {This design separates validity from difficulty}, preventing invalid problem-reference pairs from being {treated} as hard examples.

This differs from consensus-backed reward mechanisms, in which the reference answer is determined via majority voting across multiple solver runs, and problem validity is indirectly gauged by the degree of agreement among these outputs. 
In our framework (Eq.~\ref{eq:verify-score}), solver behavior only influences the setter's reward after the verifier has confirmed validity. 
The practical {\method} pipeline follows a consistent high-level workflow across tasks: (1) collect cold supervised fine-tuning (SFT) data; (2) initialize the setter via cold SFT; (3) train the setter using verifier-backed reward; (4) sample generated problem pools; (5) apply verifier and data-quality filters; (6) score accepted pairs by solver difficulty; and (7) use selected pairs for challenge evaluation or solver training.

For solver training, the verifier gate is applied at the data level. Let $\mathcal{D}_{\mathrm{V}}=\{(x,y^\star):V(x,y^\star)=1\}$ denote the verifier-accepted training pool. For each $(x,y^\star)\in\mathcal{D}_{\mathrm{V}}$, the solver receives the usual task-correctness reward, written at the problem level as
\begin{equation}
R_{\mathrm{S}}(x,y^\star)
=
\mathrm{Acc}_{S}(x,y^\star),
\qquad (x,y^\star)\in\mathcal{D}_{\mathrm{V}}.
\label{eq:solver-reward}
\end{equation}
Compared with training on unverifiable synthetic pairs, this data-level verifier gate avoids introducing invalid problem-reference pairs into the solver's reinforcement-learning signal.

The resulting procedure alternates between setter generation, verifier acceptance, solver-based difficulty scoring on accepted pairs, and downstream use of verifier-accepted pools. Algorithm~\ref{alg:vhg} in Appendix~\ref{app:implementation-details} gives the full pseudocode. In Sections~\ref{sec:integration} and~\ref{sec:math}, we instantiate the verifier $\mathrm{V}$ with a SymPy exact checker and an LLM-as-a-judge verifier, respectively.

\section{Instantiations of VHG with Hard and Soft Verifiers}
In this section, we introduce in detail how to apply the VHG framework to generate difficult and valid problems with two exemplar tasks. 
For VHG with a hard verifier, we choose the task of indefinite integral, which provides a self-contained testbed where symbolic verification is available.
We then apply VHG with a soft verifier to general math to demonstrate the broader use of VHG.

\subsection{Hard Verifier on Indefinite Integral}
\label{sec:integration}

Indefinite integral provides a perfect testbed for VHG with hard verifiers.
The problem-reference pair $(x,y^\star)$ becomes $(f,F)$, where $f$ and $F$ are the \textit{integrand} and the corresponding \textit{antiderivative}. 
This form allows us to strictly check the validity of any generated pair $(f,F)$.
In practice, we use SymPy to first validate the format of both $f$ and $F$.
Then we take the derivative $F'$ of $F$, and check whether $F'$ equals $f$.
The hard verifier for indefinite integral can be formulated as 
\begin{equation}
V_{\mathrm{int}}(f,F)
=
\mathbf{1}_{\!\left[(f,F)\in \mathcal{A}_{\mathrm{format}}\cap \mathcal{A}_{\mathrm{match}}\right]},
\label{eq:integration-score}
\end{equation}
where $\mathcal{A}_{\mathrm{format}}$ denotes the set of well-formed formulas and $\mathcal{A}_{\mathrm{match}}$ denotes the set of $(f,F)$ pairs with matched $F'$ and $f$.

\subsection{Soft Verifier on General Math}
\label{sec:math}
We extend the VHG framework to general math tasks by replacing hard verifiers with soft ones.
Specifically, for a generated problem-reference pair $(x,y^\star)$, we use LLMs to verify its validity.
General math is more open-ended than indefinite integral and therefore more exposed to reward hacking through malformed, underspecified, or superficially hard generations \citep{khalaf2025inference,helff2026gaming}.
We use explicit validation rules to make the judge's decision better grounded: before judge evaluation, a hard-coded filter rejects malformed outputs, missing or multiple final answers, trivial copies, and other failures that do not require model judgment. 
The LLM judge then checks the validity of both the problem and the answer, as well as their correspondence. 
The verifier can be formulated as
\begin{equation}
V_{\mathrm{math}}(x,y^\star)
=
\mathbf{1}_{\!\left[
(x,y^\star)\in
\mathcal{A}_{\mathrm{filter}}
\cap
\mathcal{A}_{\mathrm{LLM}}
\right]}.
\label{eq:judge-valid}
\end{equation}
where $\mathcal{A}_{\mathrm{filter}}$ denotes the hard-coded filter and $\mathcal{A}_{\mathrm{LLM}}$ denotes the LLM judge. We show our prompts for the soft verifier in Appendix~\ref{app:soft-verifier-prompts}; the rubric is task-specific and can be extended with additional checks when stronger validation criteria are available.

For both hard and soft verifier settings, to make sure setter $Q$ generates a diverse set of problems, we use existing datasets as seed sets and randomly feed seed problems as hints for $Q$ to generate harder ones. 
We initialize the setter $Q$ by supervised finetuning on a small group of examples of problem generation. The details can be found in Appendix~\ref{app:implementation-details}.

\begin{table}[t]
\caption{Representative seed-to-generated examples. The examples illustrate structural changes made by the setter in each task.}
\label{tab:seed-generated-examples}
\centering
\scriptsize
\setlength{\tabcolsep}{4pt}
\begin{tabularx}{\textwidth}{p{0.15\textwidth}XXX}
\toprule
Task & Original seed & Generated problem & Transformation illustrated \\
\midrule
\shortstack[l]{Indefinite\\integral} & Seed function $-1/x$ & $\displaystyle \int \frac{\sin(\ln |x|)-\cos(\ln |x|)}{x^2}\,dx$ & Product-style modification hides a reciprocal derivative inside a logarithmic oscillatory term. \\
\shortstack[l]{Indefinite\\integral} & Seed function $x-\arctan(x)$ & $\displaystyle \int \frac{-x^2+2x\arctan(x)}{(1+x^2)^2}\,dx$ & Rational wrapping turns a simple inverse-trigonometric seed into a quotient-rule problem. \\
\shortstack[l]{Indefinite\\integral} & Seed function $e^x-2\sqrt{x}$ & $\displaystyle \int \frac{e^x(2x+1)-1}{2\sqrt{x}}\,dx$ & Multiplication by $\sqrt{x}$ combines exponential and radical terms in the final integrand. \\
General math & How many positive divisors does $6!$ have? & How many positive divisors of $10!$ are perfect squares? & Adds a square-divisor constraint, requiring parity reasoning over prime exponents. \\
General math & What is the base-$2$ representation of $84_{10}$? & Convert the base-$5$ number $123_5$ to base $2$. & Changes direct base conversion into a two-stage conversion through decimal form. \\
General math & Fully factor $2x^2-8$. & Fully factor $12x^4-27$. & Moves from a quadratic common-factor pattern to a quartic difference-of-squares structure. \\
\bottomrule
\end{tabularx}
\end{table}

\section{Experiments}
\label{sec:experiments}
We now evaluate {\method} in both hard and soft verifier settings in two research questions:
\begin{itemize}[leftmargin=*]
\item \textbf{RQ-1}: Can {\method} generate more difficult yet valid problems as expected?
\item \textbf{RQ-2}: Can the generated problems lead to a stronger solver after RL training?
\end{itemize}

\paragraph{Experimental setup}
For both settings, we use Qwen3-4B-Base as the backbone model for the setter $Q$ and solver $S$. 
For indefinite integral, we collect a small number of problems from a college textbook and build a moderate-level difficulty seed set from it for setter training. For general math, we sample a seed pool from MATH dataset \citep{hendrycks2021math} and select an easy subset with Qwen3-4B-Base pass@1 at least 0.75 for setter RL.
To evaluate the performance of the solver, we compare \method~with several baselines, including vanilla GRPO and R-Zero, on a wide range of benchmarks, including AntiderivBench and our curated Integral Stress Test for indefinite integral, and MATH \citep{hendrycks2021math}, GSM8K \citep{cobbe2021training}, AMC, Minerva, Olympiad, AIME 2024, AIME 2025, and AIME 2026 for general math. 
A context window of \num{8192} tokens is used for the most challenging AMC and AIME benchmarks, and \num{4096} tokens is used for all other benchmarks.
We report performance using pass@1 and pass@8, computed under a fixed sampling budget for each problem.

\subsection{RQ-1: Evaluation of the Generated Problems}
\label{sec:challenge-results}

Before checking the end-to-end performance improvement of the solver, we first check whether {\method} successfully generates novel valid problems that are challenging to current LLMs.

\begin{figure}[t]
\centering
\includegraphics[width=\linewidth,trim=0bp 4bp 0bp 0bp,clip]{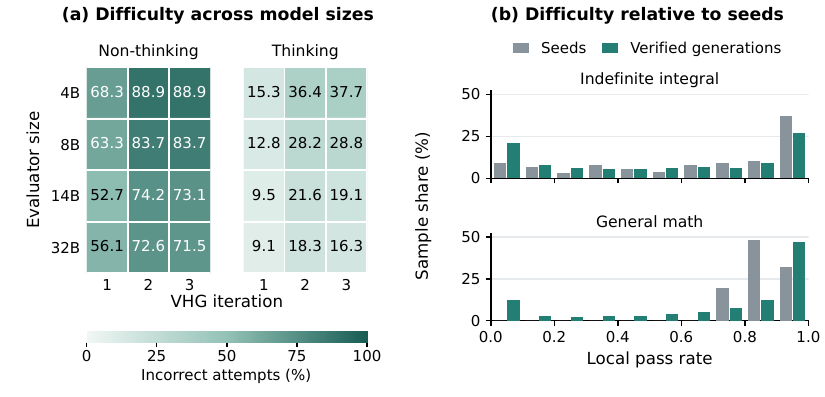}
\par\nointerlineskip
\caption{Difficulty of VHG-generated problems. (a) Problems remain challenging across model sizes. (b) Shifts in difficulty distributions from seed problems to verified generations.}
\label{fig:rq1-overview}
\label{fig:seed-valid-shift}
\label{fig:round-scaling}
\end{figure}

We use the trained setter $Q$ to generate problem-reference pairs and retain only those accepted by the verifier $V$.
Figure~\ref{fig:seed-valid-shift}(b) compares the local-solver pass-rate distribution of the setter-RL seed data with the verifier-valid generated pool after setter RL.
The generated distribution is normalized over accepted or audit-valid generated items rather than over all raw generations.
In both settings, the valid generated pool contains verifier-valid items in lower pass-rate bins that are absent or less represented in the seed data, indicating that VHG changes the difficulty profile while retaining verifier-valid problem-reference pairs.

\subsection{RQ-2: End-to-End Performance Evaluation of the Solver}
\label{sec:solver-results}

We already know that \method~can generate a diverse pool of harder valid problems, but by how much they can boost the training of the solver remains a question. 
In this part, we answer this question by comparing the solver trained with data from \method~with baseline methods.

\paragraph{Indefinite integral.}

\begin{table}[!htbp]
\caption{Integral solver-training comparison and ablation. Values are percentages.}
\label{tab:integration-comparison}
\centering
\small
\setlength{\tabcolsep}{4pt}
\resizebox{\textwidth}{!}{%
\begin{tabular}{l|rrrrrr}
\toprule
Method & Comp. Pass@1 & Comp. Pass@8 & Qual. Pass@1 & Qual. Pass@8 & Stress Test Pass@1 & Stress Test Pass@8 \\
\midrule
Qwen3-4B-Base & 28.8 & 57.0 & 52.5 & 80.0 & 43.3 & 69.8 \\
Vanilla GRPO & 38.8 & 61.2 & 66.5 & 80.3 & 60.2 & 75.4 \\
\quad w. SFT data & 42.9 & 63.5 & 66.6 & 83.3 & 57.4 & 77.3 \\
R-Zero &  &  &  &  &  &  \\
\quad Iter. 1 & 31.9 & 58.9 & 62.7 & 82.0 & 51.0 & 71.3 \\
\quad Iter. 2 & 30.7 & 56.3 & 62.8 & 81.6 & 52.9 & 71.6 \\
\quad Iter. 3 & 30.5 & 55.5 & 62.0 & 80.7 & 51.0 & 70.1 \\
{\method} (Hard) & & & & & & \\
\quad Iter.~1 & 45.4 & 66.5 & 69.4 & 83.1 & 64.7 & 81.7 \\
\quad Iter.~2 & 49.7 & 69.0 & 73.2 & 84.3 & 67.1 & \textbf{83.4} \\
\quad Iter.~3 & \textbf{51.0} & \textbf{71.4} & \textbf{74.6} & \textbf{84.4} & \textbf{67.8} & 83.1 \\
\bottomrule
\end{tabular}}
\end{table}

As shown in Table~\ref{tab:integration-comparison}, {\method} outperforms vanilla GRPO and R-Zero on all three benchmarks in both Pass@1 and Pass@8.
Pass@1 improves across iterations, exceeding R-Zero's best overall iteration by 20.3, 11.7, and 14.9 percentage points after three iterations on Competition, Qualifier, and the Integral Stress Test, respectively.
R-Zero remains below vanilla GRPO after three iterations, supporting the importance of verification in self-play.

\paragraph{General math.}
From Figure~\ref{fig:math-pass1}, we can see that {\method} outperforms R-Zero on all benchmarks except GSM8K. This is expected because {\method} aims to generate harder problems, which leads to a distribution shift from the primary-school-level problems in GSM8K.
The results indicate that a soft verifier is still beneficial in scenarios where strict verification is impossible.
Appendix~\ref{app:additional-results} reports the full benchmark and ablation results.

\subsection{Out-of-Domain Transfer}
\label{sec:ood}
\begin{figure}[!htbp]
\centering
\begin{minipage}[t]{0.48\linewidth}
\vspace{0pt}
\centering
\includegraphics[width=\linewidth,trim=0bp 19.25bp 0bp 0bp,clip]{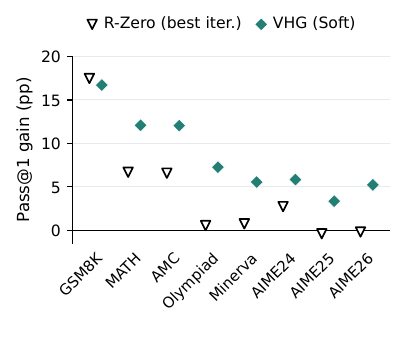}
\par\nointerlineskip
\caption{General-math pass@1 gains over the base model.}
\label{fig:math-pass1}
\end{minipage}\hfill
\begin{minipage}[t]{0.48\linewidth}
\vspace{0pt}
\centering
\includegraphics[width=\linewidth,trim=0bp 3.5bp 0bp 0bp,clip]{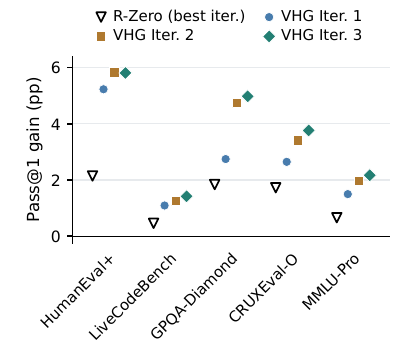}
\par\nointerlineskip
\caption{Pass@1 gains on OOD benchmarks over the base model, for R-Zero and VHG trained on the integration task.}
\label{fig:ood-gains}
\end{minipage}
\end{figure}

Figure~\ref{fig:ood-gains} shows that the integration-trained VHG solvers improve performance beyond the training domain and outperform the R-Zero baseline on all five benchmarks shown. GPQA-Diamond, CRUXEval-O, and MMLU-Pro improve with further iterations. {Full results are in Appendix~\ref{app:ood}.}

\section{Analysis on the Source of Improved Problem Generation Quality}
\label{sec:quality-analysis}

We next analyze why {\method} improves problem generation. We use two complementary views: the setter's training trajectory and the distributional characteristics of generated problems from {\method} and R-Zero. {Appendix~\ref{app:generation-analysis-details} gives additional analysis and details.}

\subsection{Setter Learning Dynamics}

\begin{figure*}[t]
\centering
\includegraphics[width=\textwidth,trim=0bp 0.25bp 0bp 0bp,clip]{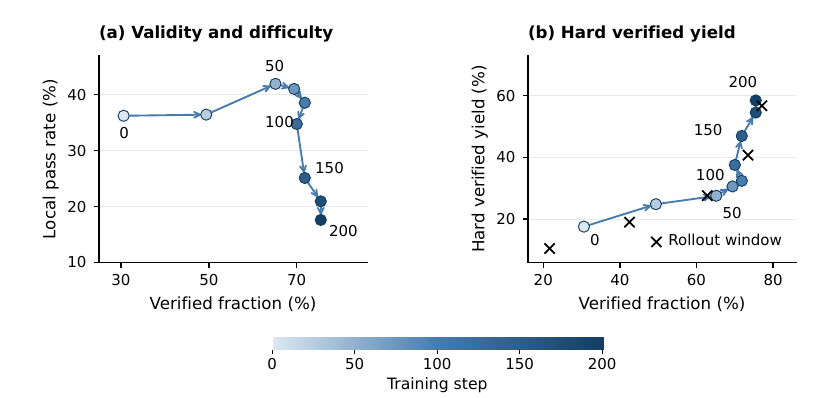}
\par\nointerlineskip
\caption{Learning trajectory of the hard-verifier setter on indefinite integral. Validity improves first; later, local pass rate decreases while hard verified yield rises. Lower local pass rate indicates harder verified problems. Hard verified yield is the fraction of all generations that are verified and have local pass rate at most 0.3. Color indicates training step.}
\label{fig:sym-hard-questioner-trajectory}
\end{figure*}

VHG aims to increase problem difficulty while maintaining validity.
We therefore examine the setter's learning dynamics to understand how these two objectives develop during training.
Figure~\ref{fig:sym-hard-questioner-trajectory} reveals a two-phase pattern: validity improves first, followed by increasing difficulty among valid generations.
From step 0 to step 50, the reference-valid rate rises from 30.6\% to 65.2\%, while the solver pass rate among valid samples slightly increases from 36.2\% to 42.0\%.
Thus, the early improvement is primarily a validity improvement, not yet a difficulty improvement.
After validity is partially established, the generated samples become substantially harder: from step 50 to step 200, the solver pass rate among valid samples falls to 17.6\%, while the reference-valid rate ends at 75.5\% after only a small intermediate decrease, and the share of all validation samples that are both valid and hard, defined here as local pass rate at most 0.3, rises from 27.5\% to 58.5\%.
The rollout-window summaries show the same direction.
This trajectory is consistent with the mechanism enabled by verifier-backed reward: once the verifier makes generated problem-reference pairs trustworthy enough to score, solver feedback can push the setter toward harder valid problems.

\subsection{Generation Distribution Analysis}

We compare VHG and R-Zero on their ability to generate difficult, valid problems.
For indefinite integration, we check validity with the hard verifier and measure difficulty using each method's local solver pass rate across three iterations.

\begin{figure*}[t]
\centering
\includegraphics[width=\textwidth,trim=0bp 3bp 0bp 0bp,clip]{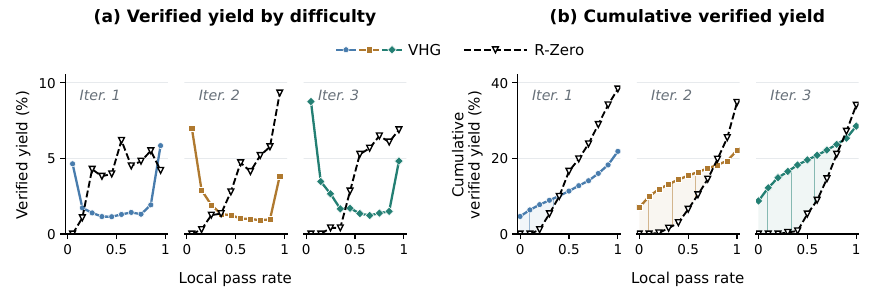}
\par\nointerlineskip
\caption{Verified yield for indefinite-integral generation across three iterations. (a) Share of each method's candidate pool that is verified within each local pass-rate bin, using bins of width 0.1 and points at bin centers. (b) Cumulative verified yield, starting with the hardest problems and progressively including easier ones, with markers at pass-rate increments of 0.1. Vertical hatching highlights where the VHG curve lies above R-Zero. {Solid and dashed curves represent VHG and R-Zero, respectively.}}
\label{fig:integration-hardness-validity}
\end{figure*}

Figure~\ref{fig:integration-hardness-validity}(a) shows that VHG produces a larger share of valid problems at the hardest end of the distribution, while R-Zero's valid generations concentrate at higher local pass rates.
Valid problems with zero local pass rate account for 4.6\%, 7.0\%, and 8.7\% of the VHG candidate pools in iterations 1--3, respectively.
Figure~\ref{fig:integration-hardness-validity}(b) further shows VHG's higher cumulative yield of valid problems in the low-pass-rate region.

This advantage follows from separating validity from solver success.
R-Zero constructs its reference answer through solver consensus and requires at least one supporting response, excluding candidates with zero consensus support.
VHG independently verifies the problem-reference pair, allowing it to retain valid problems even when all sampled solver answers are incorrect.
The verifier therefore enables the setter to pursue greater difficulty while maintaining validity.
A similar pattern in general math with the soft verifier (Figure~\ref{fig:general-math-judge-validity}) supports the same principle beyond the hard-verifier setting.

\subsection{Complexity and Diversity}
\label{sec:complexity-diversity}

Beyond difficulty and validity, we examine the complexity and diversity of VHG's generations.
For indefinite integration, we quantify complexity using expression trees: nodes represent operations, variables, and constants, while tree depth captures their nesting.
At the final checkpoints of iterations 1--3, valid generations increase in mean size from 24.16 to 31.44 and 43.87 nodes, while mean depth rises from 6.19 to 6.55 and 7.16.
Median size also rises from 24 to 28 and 41 nodes, showing that the typical generated expression becomes more complex.

In the first iteration, valid generations from hard seeds average 31.64 nodes, compared with 18.42 for easy seeds, while their solver pass rates are 7.3\% and 22.2\%, respectively.
Problems generated from hard seeds have more complex expressions and are more challenging for the solver.

\begin{figure}[t]
\centering
\includegraphics[width=\textwidth]{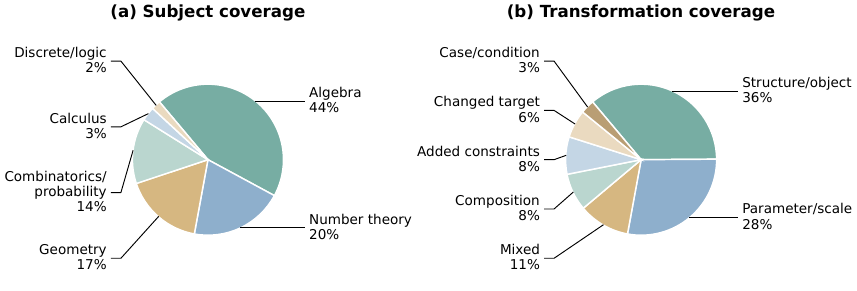}
\par\nointerlineskip
\caption{Subject and transformation coverage in 100 general-math problems generated by VHG. The distributions across (a) six subjects and (b) seven transformation types demonstrate the diversity of the generated problems.}
\label{fig:general-math-diversity-coverage}
\end{figure}

For general math, we sample 100 problems accepted by the soft verifier and use model-assisted annotation to characterize their structures, subjects, and transformation types.
The analysis finds that nearly all sampled problems have distinct structures.
Figure~\ref{fig:general-math-diversity-coverage}(a) shows that these problems span six mathematical subjects, with substantial representation from algebra, number theory, geometry, and combinatorics/probability.
Figure~\ref{fig:general-math-diversity-coverage}(b) further shows seven transformation types, including structural changes, multi-step composition, and additional constraints.
Together, the low structural repetition and the breadth of subjects and transformations show that VHG generates structurally and conceptually diverse problems.
{Appendix~\ref{app:diversity} gives detailed results and annotation procedures.}

\section{Related Work}
\label{sec:related}

\paragraph{Synthetic mathematical data.}
A large body of recent mathematical reasoning work improves solvers by expanding the supervision available around human-provided seed problems. MetaMath \citep{yu2024metamath} rewrites seed problems from multiple perspectives, WizardMath \citep{luo2024wizardmath} adapts instruction evolution and reinforcement feedback to mathematics, and MathScale \citep{tang2024mathscale} and KPMath \citep{huang2025kpmath} mine topic, concept, or key-point structure from existing problems to synthesize larger training corpora. OpenMathInstruct-2 \citep{toshniwal2025openmathinstruct2} and the AIMO-2 \citep{moshkov2025aimo2} recipe scale this direction further through generated solution traces, tool-integrated reasoning, filtering, and selection. These systems demonstrate that generated mathematical supervision can substantially improve solvers. They mostly treat the problem source as fixed or seed-driven, however, and focus on producing more diverse problems, solutions, or traces. In this paper, we make the problem-reference pair itself the object generated by the policy, so the central question shifts from how to scale supervision to when a generated hard pair is valid enough to trust.

{Symbolic and neuro-symbolic approaches construct new problems through explicit mathematical representations. For example, \citet{NEURIPS2024_29d319f7} combine symbolic solvers with mutations of formalized problems, then use an LLM to express the results in natural language. Their method controls problem complexity through predefined symbolic mutation rules. In contrast, {\method} learns how to generate challenging problems from the current solver's success rate, granting difficulty rewards only to verifier-accepted problems.}

\paragraph{Hard and verifiable problem generation.}
A closer line of work explicitly targets difficult or checkable generated problems. \citet{li2024synthesizing} abstract problems into algorithms, implement executable functions, and contextualize them into new word problems that can be checked by those functions. CHASE \citep{patel2025chase} generates challenging evaluation items bottom-up from simpler, independently verifiable subtasks. PromptCoT \citep{zhao2025promptcot} uses construction rationales to synthesize Olympiad-level problems, while SHARP \citep{wu2025sharp} and SAND-Math \citep{manem2025sandmath} add generation, filtering, and difficulty-oriented steps to obtain useful reasoning problems for RL or fine-tuning. MathSmith \citep{zhan2025mathsmith} goes further toward adaptive generation by training a reinforced policy to forge harder problems under structural and answer-consistency constraints. Our {\method} complements these efforts: it also seeks hard synthetic mathematical data, but it makes the verifier part of the setter's reward. A candidate receives a difficulty reward only after the generated problem-reference pair has been accepted, so invalid hard generations are removed from valid difficult problems, before solver failure is used as evidence of difficulty.

\paragraph{Self-play and verifier-backed reward.}
Self-evolving training has also been studied beyond math problem synthesis. SPIN \citep{chen2024spin} uses a self-play objective to improve a model from its own generations. \citet{yuan2024selfrewarding} uses the model as a judge to create iterative preference data, and DeepSeek-R1 \citep{guo2025deepseekr1} shows the strength of reinforcement learning with verifiable rewards over externally supplied prompts. The closest systems introduce generated tasks into the training loop. Absolute Zero \citep{zhao2025absolutezero} lets a model propose and solve tasks, using a code executor to validate task proposals and answers. R-Zero \citep{huang2025rzero} co-evolves a challenger and solver from zero external data; the challenger proposes problems near the solver frontier, and the solver learns from generated problem-answer data supported by consensus or pseudo-labeling. Weakness-driven and variation-based methods such as SwS \citep{liang2025sws} and SvS \citep{liang2025svs} similarly synthesize new problems from the model's current failures or correct solutions to sustain RLVR training.

Our work is closest in spirit to these setter--solver systems, especially Absolute Zero and R-Zero, but differs in where trust enters the loop. Consensus, answer extraction, and model agreement can provide useful pseudo-labels, yet they are indirect evidence of validity when the problem itself is generated. This is precisely the setting where proxy rewards can be exploited: recent work on reward hacking and verifier gaming shows that models may optimize what a verifier or reward proxy fails to enforce \citep{helff2026gaming,khalaf2025inference}. {\method} therefore makes validity a first-class gate in the reward oracle: solver failure contributes to setter reward only after a hard verifier or soft judge verifier accepts the generated problem-reference pair.

\section{Conclusion}
\label{sec:conclusion}

Hard synthetic mathematical problems should be studied through the reward oracle that selects them, not only through the generator that proposes them. Across hard-verifier integration and soft-verifier general math, {\method} checks validity before interpreting solver failure as difficulty. The exact integration setting provides the cleanest evidence: verifier-accepted generations remain difficult for stronger solvers, improve a downstream 4B solver, and expose validity gaps in consensus-generated pseudo-label pools. General math shows that the same ordering remains useful with noisier LLM-as-a-judge validation. {Together, these results show that verifier-backed self-play can generate valid, challenging problems that improve mathematical reasoning.}

\clearpage
\section*{Acknowledgments}
We would like to thank Xingyu Shen for his support on this work.
The authors have no funding or competing interests to disclose.

\bibliographystyle{plainnat}
\bibliography{references}

\begin{thebibliography}{38}
\providecommand{\natexlab}[1]{#1}
\providecommand{\url}[1]{\texttt{#1}}
\expandafter\ifx\csname urlstyle\endcsname\relax
  \providecommand{\doi}[1]{doi: #1}\else
  \providecommand{\doi}{doi: \begingroup \urlstyle{rm}\Url}\fi

\bibitem[Chen et~al.(2021)Chen, Tworek, Jun, Yuan, de~Oliveira~Pinto, Kaplan,
  Edwards, Burda, Joseph, Brockman, Ray, Puri, Krueger, Petrov, Khlaaf, Sastry,
  Mishkin, Chan, Gray, Ryder, Pavlov, Power, Kaiser, Bavarian, Winter, Tillet,
  Such, Cummings, Plappert, Chantzis, Barnes, Herbert-Voss, Guss, Nichol,
  Paino, Tezak, Tang, Babuschkin, Balaji, Jain, Saunders, Hesse, Carr, Leike,
  Achiam, Misra, Morikawa, Radford, Knight, Brundage, Murati, Mayer, Welinder,
  McGrew, Amodei, McCandlish, Sutskever, and Zaremba]{chen2021code}
Mark Chen, Jerry Tworek, Heewoo Jun, Qiming Yuan, Henrique~Ponde
  de~Oliveira~Pinto, Jared Kaplan, Harri Edwards, Yuri Burda, Nicholas Joseph,
  Greg Brockman, Alex Ray, Raul Puri, Gretchen Krueger, Michael Petrov, Heidy
  Khlaaf, Girish Sastry, Pamela Mishkin, Brooke Chan, Scott Gray, Nick Ryder,
  Mikhail Pavlov, Alethea Power, Lukasz Kaiser, Mohammad Bavarian, Clemens
  Winter, Philippe Tillet, Felipe~Petroski Such, Dave Cummings, Matthias
  Plappert, Fotios Chantzis, Elizabeth Barnes, Ariel Herbert-Voss,
  William~Hebgen Guss, Alex Nichol, Alex Paino, Nikolas Tezak, Jie Tang, Igor
  Babuschkin, Suchir Balaji, Shantanu Jain, William Saunders, Christopher
  Hesse, Andrew~N. Carr, Jan Leike, Josh Achiam, Vedant Misra, Evan Morikawa,
  Alec Radford, Matthew Knight, Miles Brundage, Mira Murati, Katie Mayer, Peter
  Welinder, Bob McGrew, Dario Amodei, Sam McCandlish, Ilya Sutskever, and
  Wojciech Zaremba.
\newblock {Evaluating Large Language Models Trained on Code}, 2021.
\newblock URL \url{https://arxiv.org/abs/2107.03374}.
\newblock arXiv:2107.03374.

\bibitem[Chen et~al.(2024)Chen, Deng, Yuan, Ji, and Gu]{chen2024spin}
Zixiang Chen, Yihe Deng, Huizhuo Yuan, Kaixuan Ji, and Quanquan Gu.
\newblock {Self-Play Fine-Tuning Converts Weak Language Models to Strong
  Language Models}.
\newblock In Ruslan Salakhutdinov, Zico Kolter, Katherine Heller, Adrian
  Weller, Nuria Oliver, Jonathan Scarlett, and Felix Berkenkamp, editors,
  \emph{Proceedings of the 41st International Conference on Machine Learning},
  volume 235 of \emph{Proceedings of Machine Learning Research}, pages
  6621--6642. PMLR, 21--27 Jul 2024.
\newblock URL \url{https://proceedings.mlr.press/v235/chen24j.html}.

\bibitem[Cobbe et~al.(2021)Cobbe, Kosaraju, Bavarian, Chen, Jun, Kaiser,
  Plappert, Tworek, Hilton, Nakano, Hesse, and Schulman]{cobbe2021training}
Karl Cobbe, Vineet Kosaraju, Mohammad Bavarian, Mark Chen, Heewoo Jun, Lukasz
  Kaiser, Matthias Plappert, Jerry Tworek, Jacob Hilton, Reiichiro Nakano,
  Christopher Hesse, and John Schulman.
\newblock {Training Verifiers to Solve Math Word Problems}, 2021.
\newblock URL \url{https://arxiv.org/abs/2110.14168}.
\newblock arXiv:2110.14168.

\bibitem[Gao et~al.(2026)Gao, Kim, Sun, Joachims, Wang, Pang, and
  Tan]{gao2025promptcurriculum}
Zhaolin Gao, Joongwon Kim, Wen Sun, Thorsten Joachims, Sid Wang,
  Richard~Yuanzhe Pang, and Liang Tan.
\newblock {Prompt Curriculum Learning for Efficient {LLM} Post-Training}.
\newblock In C.~Vondrick, B.~Hariharan, C.~Raffel, L.~Pinto, D.~Yang, and
  A.~Faust, editors, \emph{International Conference on Learning
  Representations}, volume 2026, pages 58614--58652, 2026.
\newblock URL
  \url{https://proceedings.iclr.cc/paper_files/paper/2026/file/601b42cebb3650ddb075770e2a85f428-Paper-Conference.pdf}.

\bibitem[Gou et~al.(2024)Gou, Shao, Gong, shen, Yang, Huang, Duan, and
  Chen]{gou2024tora}
Zhibin Gou, Zhihong Shao, Yeyun Gong, yelong shen, Yujiu Yang, Minlie Huang,
  Nan Duan, and Weizhu Chen.
\newblock {To{RA}: A Tool-Integrated Reasoning Agent for Mathematical Problem
  Solving}.
\newblock In B.~Kim, Y.~Yue, S.~Chaudhuri, K.~Fragkiadaki, M.~Khan, and Y.~Sun,
  editors, \emph{International Conference on Learning Representations}, volume
  2024, pages 48362--48395, 2024.
\newblock URL
  \url{https://proceedings.iclr.cc/paper_files/paper/2024/file/d3cf1559a8795eb1ed2b3ad52409ac7d-Paper-Conference.pdf}.

\bibitem[Gu et~al.(2024)Gu, Roziere, Leather, Solar-Lezama, Synnaeve, and
  Wang]{gu2024cruxeval}
Alex Gu, Baptiste Roziere, Hugh~James Leather, Armando Solar-Lezama, Gabriel
  Synnaeve, and Sida Wang.
\newblock {{CRUXE}val: A Benchmark for Code Reasoning, Understanding and
  Execution}.
\newblock In Ruslan Salakhutdinov, Zico Kolter, Katherine Heller, Adrian
  Weller, Nuria Oliver, Jonathan Scarlett, and Felix Berkenkamp, editors,
  \emph{Proceedings of the 41st International Conference on Machine Learning},
  volume 235 of \emph{Proceedings of Machine Learning Research}, pages
  16568--16621. PMLR, 21--27 Jul 2024.
\newblock URL \url{https://proceedings.mlr.press/v235/gu24c.html}.

\bibitem[Guo et~al.(2025)Guo, Yang, Zhang, Song, Wang, Zhu, Xu, Zhang, Ma, Bi,
  Zhang, Yu, Wu, Wu, Gou, Shao, Li, Gao, Liu, Xue, Wang, Wu, Feng, Lu, Zhao,
  Deng, Ruan, Dai, Chen, Ji, Li, Lin, Dai, Luo, Hao, Chen, Li, Zhang, Xu, Ding,
  Gao, Qu, Li, Guo, Li, Chen, Yuan, Tu, Qiu, Li, Cai, Ni, Liang, Chen, Dong,
  Hu, You, Gao, Guan, Huang, Yu, Wang, Zhang, Zhao, Wang, Zhang, Xu, Xia,
  Zhang, Zhang, Tang, Zhou, Li, Wang, Li, Tian, Huang, Zhang, Wang, Chen, Du,
  Ge, Zhang, Pan, Wang, Chen, Jin, Chen, Lu, Zhou, Chen, Ye, Wang, Yu, Zhou,
  Pan, Li, Zhou, Wu, Yun, Pei, Sun, Wang, Zeng, Liu, Liang, Gao, Yu, Zhang,
  Xiao, An, Liu, Wang, Chen, Nie, Cheng, Liu, Xie, Liu, Yang, Li, Su, Lin, Li,
  Jin, Shen, Chen, Sun, Wang, Song, Zhou, Wang, Shan, Li, Wang, Wei, Zhang, Xu,
  Li, Zhao, Sun, Wang, Yu, Zhang, Shi, Xiong, He, Piao, Wang, Tan, Ma, Liu,
  Guo, Ou, Wang, Gong, Zou, He, Xiong, Luo, You, Liu, Zhou, Zhu, Huang, Li,
  Zheng, Zhu, Ma, Tang, Zha, Yan, Ren, Ren, Sha, Fu, Xu, Xie, Zhang, Hao, Ma,
  Yan, Wu, Gu, Zhu, Liu, Li, Xie, Song, Pan, Huang, Xu, Zhang, and
  Zhang]{guo2025deepseekr1}
Daya Guo, Dejian Yang, Haowei Zhang, Junxiao Song, Peiyi Wang, Qihao Zhu,
  Runxin Xu, Ruoyu Zhang, Shirong Ma, Xiao Bi, Xiaokang Zhang, Xingkai Yu,
  Yu~Wu, Z.~F. Wu, Zhibin Gou, Zhihong Shao, Zhuoshu Li, Ziyi Gao, Aixin Liu,
  Bing Xue, Bingxuan Wang, Bochao Wu, Bei Feng, Chengda Lu, Chenggang Zhao,
  Chengqi Deng, Chong Ruan, Damai Dai, Deli Chen, Dongjie Ji, Erhang Li,
  Fangyun Lin, Fucong Dai, Fuli Luo, Guangbo Hao, Guanting Chen, Guowei Li,
  H.~Zhang, Hanwei Xu, Honghui Ding, Huazuo Gao, Hui Qu, Hui Li, Jianzhong Guo,
  Jiashi Li, Jingchang Chen, Jingyang Yuan, Jinhao Tu, Junjie Qiu, Junlong Li,
  J.~L. Cai, Jiaqi Ni, Jian Liang, Jin Chen, Kai Dong, Kai Hu, Kaichao You,
  Kaige Gao, Kang Guan, Kexin Huang, Kuai Yu, Lean Wang, Lecong Zhang, Liang
  Zhao, Litong Wang, Liyue Zhang, Lei Xu, Leyi Xia, Mingchuan Zhang, Minghua
  Zhang, Minghui Tang, Mingxu Zhou, Meng Li, Miaojun Wang, Mingming Li, Ning
  Tian, Panpan Huang, Peng Zhang, Qiancheng Wang, Qinyu Chen, Qiushi Du, Ruiqi
  Ge, Ruisong Zhang, Ruizhe Pan, Runji Wang, R.~J. Chen, R.~L. Jin, Ruyi Chen,
  Shanghao Lu, Shangyan Zhou, Shanhuang Chen, Shengfeng Ye, Shiyu Wang,
  Shuiping Yu, Shunfeng Zhou, Shuting Pan, S.~S. Li, Shuang Zhou, Shaoqing Wu,
  Tao Yun, Tian Pei, Tianyu Sun, T.~Wang, Wangding Zeng, Wen Liu, Wenfeng
  Liang, Wenjun Gao, Wenqin Yu, Wentao Zhang, W.~L. Xiao, Wei An, Xiaodong Liu,
  Xiaohan Wang, Xiaokang Chen, Xiaotao Nie, Xin Cheng, Xin Liu, Xin Xie,
  Xingchao Liu, Xinyu Yang, Xinyuan Li, Xuecheng Su, Xuheng Lin, X.~Q. Li,
  Xiangyue Jin, Xiaojin Shen, Xiaosha Chen, Xiaowen Sun, Xiaoxiang Wang, Xinnan
  Song, Xinyi Zhou, Xianzu Wang, Xinxia Shan, Y.~K. Li, Y.~Q. Wang, Y.~X. Wei,
  Yang Zhang, Yanhong Xu, Yao Li, Yao Zhao, Yaofeng Sun, Yaohui Wang, Yi~Yu,
  Yichao Zhang, Yifan Shi, Yiliang Xiong, Ying He, Yishi Piao, Yisong Wang,
  Yixuan Tan, Yiyang Ma, Yiyuan Liu, Yongqiang Guo, Yuan Ou, Yuduan Wang, Yue
  Gong, Yuheng Zou, Yujia He, Yunfan Xiong, Yuxiang Luo, Yuxiang You, Yuxuan
  Liu, Yuyang Zhou, Y.~X. Zhu, Yanping Huang, Yaohui Li, Yi~Zheng, Yuchen Zhu,
  Yunxian Ma, Ying Tang, Yukun Zha, Yuting Yan, Z.~Z. Ren, Zehui Ren, Zhangli
  Sha, Zhe Fu, Zhean Xu, Zhenda Xie, Zhengyan Zhang, Zhewen Hao, Zhicheng Ma,
  Zhigang Yan, Zhiyu Wu, Zihui Gu, Zijia Zhu, Zijun Liu, Zilin Li, Ziwei Xie,
  Ziyang Song, Zizheng Pan, Zhen Huang, Zhipeng Xu, Zhongyu Zhang, and Zhen
  Zhang.
\newblock {DeepSeek-R1 incentivizes reasoning in LLMs through reinforcement
  learning}.
\newblock \emph{Nature}, 645\penalty0 (8081):\penalty0 633--638, Sep 2025.
\newblock ISSN 1476-4687.
\newblock \doi{10.1038/s41586-025-09422-z}.
\newblock URL \url{https://doi.org/10.1038/s41586-025-09422-z}.

\bibitem[Helff et~al.(2026)Helff, Delfosse, Steinmann, H{\"a}rle, Shindo,
  Schramowski, Stammer, Kersting, and Friedrich]{helff2026gaming}
Lukas Helff, Quentin Delfosse, David Steinmann, Ruben H{\"a}rle, Hikaru Shindo,
  Patrick Schramowski, Wolfgang Stammer, Kristian Kersting, and Felix
  Friedrich.
\newblock {{LLM}s Gaming Verifiers: {RLVR} can Lead to Reward Hacking}.
\newblock In \emph{ICLR 2026 Workshop on Logical Reasoning of Large Language
  Models}, 2026.
\newblock URL \url{https://openreview.net/forum?id=4B3WfRNqe3}.

\bibitem[Hendrycks et~al.(2021)Hendrycks, Burns, Kadavath, Arora, Basart, Tang,
  Song, and Steinhardt]{hendrycks2021math}
Dan Hendrycks, Collin Burns, Saurav Kadavath, Akul Arora, Steven Basart, Eric
  Tang, Dawn Song, and Jacob Steinhardt.
\newblock {Measuring Mathematical Problem Solving With the MATH Dataset}.
\newblock In J.~Vanschoren and S.~Yeung, editors, \emph{Proceedings of the
  Neural Information Processing Systems Track on Datasets and Benchmarks},
  volume~1, 2021.
\newblock URL
  \url{https://datasets-benchmarks-proceedings.neurips.cc/paper_files/paper/2021/file/be83ab3ecd0db773eb2dc1b0a17836a1-Paper-round2.pdf}.

\bibitem[Huang et~al.(2026)Huang, Yu, Wang, Zhang, Li, Li, Huang, Mi, and
  Yu]{huang2025rzero}
Chengsong Huang, Wenhao Yu, Xiaoyang Wang, Hongming Zhang, Zongxia Li, Ruosen
  Li, Jiaxin Huang, Haitao Mi, and Dong Yu.
\newblock {R-Zero: Self-Evolving Reasoning {LLM} from Zero Data}.
\newblock In C.~Vondrick, B.~Hariharan, C.~Raffel, L.~Pinto, D.~Yang, and
  A.~Faust, editors, \emph{International Conference on Learning
  Representations}, volume 2026, pages 130770--130790, 2026.
\newblock URL
  \url{https://proceedings.iclr.cc/paper_files/paper/2026/file/d49b9aacebda61051166335af6fd3061-Paper-Conference.pdf}.

\bibitem[Huang et~al.(2025)Huang, Liu, Gong, Gou, Shen, Duan, and
  Chen]{huang2025kpmath}
Yiming Huang, Xiao Liu, Yeyun Gong, Zhibin Gou, Yelong Shen, Nan Duan, and
  Weizhu Chen.
\newblock {Key-Point-Driven Data Synthesis with Its Enhancement on Mathematical
  Reasoning}.
\newblock \emph{Proceedings of the AAAI Conference on Artificial Intelligence},
  39\penalty0 (23):\penalty0 24176--24184, Apr 2025.
\newblock ISSN 2159-5399.
\newblock \doi{10.1609/aaai.v39i23.34593}.
\newblock URL \url{http://dx.doi.org/10.1609/aaai.v39i23.34593}.

\bibitem[Hubert et~al.(2026)Hubert, Mehta, Sartran, Horv{\'a}th,
  {\v{Z}}u{\v{z}}i{\'{c}}, Wieser, Huang, Schrittwieser, Schroecker, Masoom,
  Bertolli, Zahavy, Mandhane, Yung, Beloshapka, Ibarz, Veeriah, Yu, Nash,
  Lezeau, Mercuri, S{\"o}nne, Mehta, Davies, Zheng, Pedregosa, Li, von Glehn,
  Rowland, Albanie, Velingker, Schmitt, Lockhart, Hughes, Michalewski,
  Sonnerat, Hassabis, Kohli, and Silver]{hubert2026alphaproof}
Thomas Hubert, Rishi Mehta, Laurent Sartran, Mikl{\'o}s~Z. Horv{\'a}th, Goran
  {\v{Z}}u{\v{z}}i{\'{c}}, Eric Wieser, Aja Huang, Julian Schrittwieser,
  Yannick Schroecker, Hussain Masoom, Ottavia Bertolli, Tom Zahavy, Amol
  Mandhane, Jessica Yung, Iuliya Beloshapka, Borja Ibarz, Vivek Veeriah, Lei
  Yu, Oliver Nash, Paul Lezeau, Salvatore Mercuri, Calle S{\"o}nne, Bhavik
  Mehta, Alex Davies, Daniel Zheng, Fabian Pedregosa, Yin Li, Ingrid von Glehn,
  Mark Rowland, Samuel Albanie, Ameya Velingker, Simon Schmitt, Edward
  Lockhart, Edward Hughes, Henryk Michalewski, Nicolas Sonnerat, Demis
  Hassabis, Pushmeet Kohli, and David Silver.
\newblock {Olympiad-level formal mathematical reasoning with reinforcement
  learning}.
\newblock \emph{Nature}, 651\penalty0 (8106):\penalty0 607--613, Mar 2026.
\newblock ISSN 1476-4687.
\newblock \doi{10.1038/s41586-025-09833-y}.
\newblock URL \url{https://doi.org/10.1038/s41586-025-09833-y}.

\bibitem[Jain et~al.(2025)Jain, Han, Gu, Li, Yan, Zhang, Wang, Solar-Lezama,
  Sen, and Stoica]{jain2024livecodebench}
Naman Jain, Han, Alex Gu, Wen-Ding Li, Fanjia Yan, Tianjun Zhang, Sida Wang,
  Armando Solar-Lezama, Koushik Sen, and Ion Stoica.
\newblock {LiveCodeBench: Holistic and Contamination Free Evaluation of Large
  Language Models for Code}.
\newblock In Y.~Yue, A.~Garg, N.~Peng, F.~Sha, and R.~Yu, editors,
  \emph{International Conference on Learning Representations}, volume 2025,
  pages 58791--58831, 2025.
\newblock URL
  \url{https://proceedings.iclr.cc/paper_files/paper/2025/file/94074dd5a072d28ff75a76dabed43767-Paper-Conference.pdf}.

\bibitem[Khalaf et~al.(2025)Khalaf, Mayrink~Verdun, Oesterling, Lakkaraju, and
  Calmon]{khalaf2025inference}
Hadi Khalaf, Claudio Mayrink~Verdun, Alex Oesterling, Himabindu Lakkaraju, and
  Flavio Calmon.
\newblock {Inference-Time Reward Hacking in Large Language Models}.
\newblock In D.~Belgrave, C.~Zhang, H.~Lin, R.~Pascanu, P.~Koniusz,
  M.~Ghassemi, and N.~Chen, editors, \emph{Advances in Neural Information
  Processing Systems}, volume 38, Main Conference, pages 61720--61760. Curran
  Associates, Inc., 2025.
\newblock \doi{10.52202/085713-2064}.
\newblock URL
  \url{https://proceedings.neurips.cc/paper_files/paper/2025/file/590a0cc0306c1c63e2d66a51a407718f-Paper-Conference.pdf}.

\bibitem[Li et~al.(2024{\natexlab{a}})Li, He, and Liu]{li2024synthesizing}
Xuefeng Li, Yanheng He, and Pengfei Liu.
\newblock {Synthesizing Verified Mathematical Problems}.
\newblock In \emph{The 4th Workshop on Mathematical Reasoning and AI at
  NeurIPS'24}, 2024{\natexlab{a}}.
\newblock URL \url{https://openreview.net/forum?id=L5US093OwO}.

\bibitem[Li et~al.(2024{\natexlab{b}})Li, Zhou, Yao, Li, Cao, Yang, Zhang, and
  Ma]{NEURIPS2024_29d319f7}
Zenan Li, Zhi Zhou, Yuan Yao, Yu-Feng Li, Chun Cao, Fan Yang, Xian Zhang, and
  Xiaoxing Ma.
\newblock Neuro-symbolic data generation for math reasoning.
\newblock In A.~Globerson, L.~Mackey, D.~Belgrave, A.~Fan, U.~Paquet,
  J.~Tomczak, and C.~Zhang, editors, \emph{Advances in Neural Information
  Processing Systems}, volume~37, pages 23488--23515. Curran Associates, Inc.,
  2024{\natexlab{b}}.
\newblock \doi{10.52202/079017-0740}.
\newblock URL
  \url{https://proceedings.neurips.cc/paper_files/paper/2024/file/29d319f7c1513c9ecd81d3a6e9632a6e-Paper-Conference.pdf}.

\bibitem[Liang et~al.(2025)Liang, Li, Gong, Wang, Zhang, shen, Wu, and
  Chen]{liang2025sws}
Xiao Liang, Zhong-Zhi Li, Yeyun Gong, Yang Wang, Hengyuan Zhang, yelong shen,
  Ying~Nian Wu, and Weizhu Chen.
\newblock {SwS: Self-aware Weakness-driven Problem Synthesis in Reinforcement
  Learning for {LLM} Reasoning}.
\newblock In D.~Belgrave, C.~Zhang, H.~Lin, R.~Pascanu, P.~Koniusz,
  M.~Ghassemi, and N.~Chen, editors, \emph{Advances in Neural Information
  Processing Systems}, volume 38, Main Conference, pages 56801--56839. Curran
  Associates, Inc., 2025.
\newblock \doi{10.52202/085713-1901}.
\newblock URL
  \url{https://proceedings.neurips.cc/paper_files/paper/2025/file/522cc00c26345fa54815ed6262f13a63-Paper-Conference.pdf}.

\bibitem[Liang et~al.(2026)Liang, Li, Gong, shen, Wu, Guo, and
  Chen]{liang2025svs}
Xiao Liang, Zhong-Zhi Li, Yeyun Gong, yelong shen, Yingnian Wu, Zhijiang Guo,
  and Weizhu Chen.
\newblock {Beyond Pass@ 1: Self-Play with Variational Problem Synthesis
  Sustains {RLVR}}.
\newblock In C.~Vondrick, B.~Hariharan, C.~Raffel, L.~Pinto, D.~Yang, and
  A.~Faust, editors, \emph{International Conference on Learning
  Representations}, volume 2026, pages 79443--79491, 2026.
\newblock URL
  \url{https://proceedings.iclr.cc/paper_files/paper/2026/file/805cd624d38e39ca061e8fd431e20825-Paper-Conference.pdf}.

\bibitem[Liu et~al.(2023)Liu, Xia, Wang, and ZHANG]{liu2023evalplus}
Jiawei Liu, Chunqiu~Steven Xia, Yuyao Wang, and LINGMING ZHANG.
\newblock {Is Your Code Generated by Chat{GPT} Really Correct? Rigorous
  Evaluation of Large Language Models for Code Generation}.
\newblock In A.~Oh, T.~Naumann, A.~Globerson, K.~Saenko, M.~Hardt, and
  S.~Levine, editors, \emph{Advances in Neural Information Processing Systems},
  volume~36, pages 21558--21572. Curran Associates, Inc., 2023.
\newblock \doi{10.52202/075280-0943}.
\newblock URL
  \url{https://proceedings.neurips.cc/paper_files/paper/2023/file/43e9d647ccd3e4b7b5baab53f0368686-Paper-Conference.pdf}.

\bibitem[Luo et~al.(2025)Luo, Sun, Xu, Zhao, Lou, Tao, Geng, Lin, Chen, Tang,
  and Zhang]{luo2024wizardmath}
Haipeng Luo, Qingfeng Sun, Can Xu, Pu~Zhao, Jian-Guang Lou, Chongyang Tao,
  Xiubo Geng, Qingwei Lin, Shifeng Chen, Yansong Tang, and Dongmei Zhang.
\newblock {WizardMath: Empowering Mathematical Reasoning for Large Language
  Models via Reinforced Evol-Instruct}.
\newblock In Y.~Yue, A.~Garg, N.~Peng, F.~Sha, and R.~Yu, editors,
  \emph{International Conference on Learning Representations}, volume 2025,
  pages 49573--49609, 2025.
\newblock URL
  \url{https://proceedings.iclr.cc/paper_files/paper/2025/file/7c04aea54c2a60a632a47bd451cd2849-Paper-Conference.pdf}.

\bibitem[Manem et~al.(2025)Manem, Brahma, Mishra, Liu, and
  Barsoum]{manem2025sandmath}
Chaitanya Manem, Pratik~Prabhanjan Brahma, Prakamya Mishra, Zicheng Liu, and
  Emad Barsoum.
\newblock {SAND-Math: Using LLMs to Generate Novel, Difficult and Useful
  Mathematics Questions and Answers}.
\newblock In \emph{The 5th Workshop on Mathematical Reasoning and AI at NeurIPS
  2025}, 2025.
\newblock URL \url{https://mathai2025.github.io/papers/}.

\bibitem[Moshkov et~al.(2025)Moshkov, Hanley, Sorokin, Toshniwal, Henkel,
  Schifferer, Du, and Gitman]{moshkov2025aimo2}
Ivan Moshkov, Darragh Hanley, Ivan Sorokin, Shubham Toshniwal, Christof Henkel,
  Benedikt Schifferer, Wei Du, and Igor Gitman.
\newblock {AIMO-2 Winning Solution: Building State-of-the-Art Mathematical
  Reasoning Models with OpenMathReasoning dataset}, 2025.
\newblock URL \url{https://arxiv.org/abs/2504.16891}.
\newblock arXiv:2504.16891.

\bibitem[{OpenAI}(2024)]{openai2024learning}
{OpenAI}.
\newblock {Learning to reason with LLMs}.
\newblock \url{https://openai.com/index/learning-to-reason-with-llms/},
  September 2024.

\bibitem[Patel et~al.(2025)Patel, Reddy, and Bahdanau]{patel2025chase}
Arkil Patel, Siva Reddy, and Dzmitry Bahdanau.
\newblock {How to Get Your {LLM} to Generate Challenging Problems for
  Evaluation}.
\newblock In \emph{NeurIPS 2025 Workshop on Evaluating the Evolving LLM
  Lifecycle: Benchmarks, Emergent Abilities, and Scaling}, 2025.
\newblock URL \url{https://openreview.net/forum?id=AQm9quyPHU}.

\bibitem[Pei et~al.(2025)Pei, Wu, Pan, Li, Lin, Ming, Gao, He, and
  Yan]{pei2025mathfusion}
Qizhi Pei, Lijun Wu, Zhuoshi Pan, Yu~Li, Honglin Lin, Chenlin Ming, Xin Gao,
  Conghui He, and Rui Yan.
\newblock {{M}ath{F}usion: Enhancing Mathematical Problem-solving of {LLM}
  through Instruction Fusion}.
\newblock In Wanxiang Che, Joyce Nabende, Ekaterina Shutova, and Mohammad~Taher
  Pilehvar, editors, \emph{Proceedings of the 63rd Annual Meeting of the
  {A}ssociation for {C}omputational {L}inguistics (Volume 1: Long Papers)},
  pages 7400--7420, Vienna, Austria, jul 2025. Association for Computational
  Linguistics.
\newblock ISBN 979-8-89176-251-0.
\newblock \doi{10.18653/v1/2025.acl-long.367}.
\newblock URL \url{https://aclanthology.org/2025.acl-long.367/}.

\bibitem[Rein et~al.(2024)Rein, Hou, Stickland, Petty, Pang, Dirani, Michael,
  and Bowman]{rein2024gpqa}
David Rein, Betty~Li Hou, Asa~Cooper Stickland, Jackson Petty, Richard~Yuanzhe
  Pang, Julien Dirani, Julian Michael, and Samuel~R. Bowman.
\newblock {{GPQA}: A Graduate-Level Google-Proof Q\&A Benchmark}.
\newblock In \emph{First Conference on Language Modeling}, 2024.
\newblock URL \url{https://openreview.net/forum?id=Ti67584b98}.

\bibitem[Shao et~al.(2024)Shao, Wang, Zhu, Xu, Song, Bi, Zhang, Zhang, Li, Wu,
  and Guo]{shao2024deepseekmath}
Zhihong Shao, Peiyi Wang, Qihao Zhu, Runxin Xu, Junxiao Song, Xiao Bi, Haowei
  Zhang, Mingchuan Zhang, Y.~K. Li, Y.~Wu, and Daya Guo.
\newblock {DeepSeekMath: Pushing the Limits of Mathematical Reasoning in Open
  Language Models}, 2024.
\newblock URL \url{https://arxiv.org/abs/2402.03300}.
\newblock arXiv:2402.03300.

\bibitem[Tang et~al.(2024)Tang, Zhang, Wang, and Wei]{tang2024mathscale}
Zhengyang Tang, Xingxing Zhang, Benyou Wang, and Furu Wei.
\newblock {{M}ath{S}cale: Scaling Instruction Tuning for Mathematical
  Reasoning}.
\newblock In Ruslan Salakhutdinov, Zico Kolter, Katherine Heller, Adrian
  Weller, Nuria Oliver, Jonathan Scarlett, and Felix Berkenkamp, editors,
  \emph{Proceedings of the 41st International Conference on Machine Learning},
  volume 235 of \emph{Proceedings of Machine Learning Research}, pages
  47885--47900. PMLR, 21--27 Jul 2024.
\newblock URL \url{https://proceedings.mlr.press/v235/tang24k.html}.

\bibitem[Toshniwal et~al.(2025)Toshniwal, Du, Moshkov, Kisacanin, Ayrapetyan,
  and Gitman]{toshniwal2025openmathinstruct2}
Shubham Toshniwal, Wei Du, Ivan Moshkov, Branislav Kisacanin, Alexan
  Ayrapetyan, and Igor Gitman.
\newblock {OpenMathInstruct-2: Accelerating {AI} for Math with Massive
  Open-Source Instruction Data}.
\newblock In Y.~Yue, A.~Garg, N.~Peng, F.~Sha, and R.~Yu, editors,
  \emph{International Conference on Learning Representations}, volume 2025,
  pages 19243--19275, 2025.
\newblock URL
  \url{https://proceedings.iclr.cc/paper_files/paper/2025/file/302ce0673c00aee2cf84bb43d0117553-Paper-Conference.pdf}.

\bibitem[Trinh et~al.(2024)Trinh, Wu, Le, He, and
  Luong]{trinh2024alphageometry}
Trieu~H. Trinh, Yuhuai Wu, Quoc~V. Le, He~He, and Thang Luong.
\newblock {Solving olympiad geometry without human demonstrations}.
\newblock \emph{Nature}, 625\penalty0 (7995):\penalty0 476--482, Jan 2024.
\newblock ISSN 1476-4687.
\newblock \doi{10.1038/s41586-023-06747-5}.
\newblock URL \url{https://doi.org/10.1038/s41586-023-06747-5}.

\bibitem[Wang et~al.(2024)Wang, Ma, Zhang, Ni, Chandra, Guo, Ren, Arulraj, He,
  Jiang, Li, Ku, Wang, Zhuang, Fan, Yue, and Chen]{wang2024mmlupro}
Yubo Wang, Xueguang Ma, Ge~Zhang, Yuansheng Ni, Abhranil Chandra, Shiguang Guo,
  Weiming Ren, Aaran Arulraj, Xuan He, Ziyan Jiang, Tianle Li, Max Ku, Kai
  Wang, Alex Zhuang, Rongqi Fan, Xiang Yue, and Wenhu Chen.
\newblock {{MMLU}-Pro: A More Robust and Challenging Multi-Task Language
  Understanding Benchmark}.
\newblock In A.~Globerson, L.~Mackey, D.~Belgrave, A.~Fan, U.~Paquet,
  J.~Tomczak, and C.~Zhang, editors, \emph{Advances in Neural Information
  Processing Systems}, volume~37, pages 95266--95290. Curran Associates, Inc.,
  2024.
\newblock \doi{10.52202/079017-3018}.
\newblock URL
  \url{https://proceedings.neurips.cc/paper_files/paper/2024/file/ad236edc564f3e3156e1b2feafb99a24-Paper-Datasets_and_Benchmarks_Track.pdf}.

\bibitem[Wu et~al.(2025)Wu, Zhang, Wen, Zhang, Ren, Shi, Chen, Zhao, Wang, Han,
  Tang, Jin, Cui, and Zhou]{wu2025sharp}
Xiong~Jun Wu, Zhenduo Zhang, ZuJie Wen, Zhiqiang Zhang, Wang Ren, Lei Shi, Cai
  Chen, Deng Zhao, Qing Wang, Xudong Han, Chengfu Tang, Dingnan Jin, Qing Cui,
  and Jun Zhou.
\newblock {SHARP: Synthesizing High-quality Aligned Reasoning Problems for
  Large Reasoning Models Reinforcement Learning}, 2025.
\newblock URL \url{https://arxiv.org/abs/2505.14147}.
\newblock arXiv:2505.14147.

\bibitem[Yu et~al.(2024)Yu, JIANG, Shi, YU, Liu, Zhang, Kwok, Li, Weller, and
  Liu]{yu2024metamath}
Longhui Yu, Weisen JIANG, Han Shi, Jincheng YU, Zhengying Liu, Yu~Zhang, James
  Kwok, Zhenguo Li, Adrian Weller, and Weiyang Liu.
\newblock {MetaMath: Bootstrap Your Own Mathematical Questions for Large
  Language Models}.
\newblock In B.~Kim, Y.~Yue, S.~Chaudhuri, K.~Fragkiadaki, M.~Khan, and Y.~Sun,
  editors, \emph{International Conference on Learning Representations}, volume
  2024, pages 45040--45061, 2024.
\newblock URL
  \url{https://proceedings.iclr.cc/paper_files/paper/2024/file/c400474e8a36d0812fdee52739288b12-Paper-Conference.pdf}.

\bibitem[Yuan et~al.(2024)Yuan, Pang, Cho, Li, Sukhbaatar, Xu, and
  Weston]{yuan2024selfrewarding}
Weizhe Yuan, Richard~Yuanzhe Pang, Kyunghyun Cho, Xian Li, Sainbayar
  Sukhbaatar, Jing Xu, and Jason~E Weston.
\newblock {Self-Rewarding Language Models}.
\newblock In Ruslan Salakhutdinov, Zico Kolter, Katherine Heller, Adrian
  Weller, Nuria Oliver, Jonathan Scarlett, and Felix Berkenkamp, editors,
  \emph{Proceedings of the 41st International Conference on Machine Learning},
  volume 235 of \emph{Proceedings of Machine Learning Research}, pages
  57905--57923. PMLR, 21--27 Jul 2024.
\newblock URL \url{https://proceedings.mlr.press/v235/yuan24d.html}.

\bibitem[Zhan et~al.(2026)Zhan, Lai, Lu, Lin, Yang, and Tan]{zhan2025mathsmith}
Shaoxiong Zhan, Yanlin Lai, Ziyu Lu, Dahua Lin, Ziqing Yang, and Fei Tan.
\newblock {MathSmith: Towards Extremely Hard Mathematical Reasoning by Forging
  Synthetic Problems with a Reinforced Policy}.
\newblock \emph{Proceedings of the AAAI Conference on Artificial Intelligence},
  40\penalty0 (41):\penalty0 34602--34610, Mar 2026.
\newblock ISSN 2159-5399.
\newblock \doi{10.1609/aaai.v40i41.40760}.
\newblock URL \url{http://dx.doi.org/10.1609/aaai.v40i41.40760}.

\bibitem[Zhang et~al.(2025)Zhang, Li, Long, Zhang, Lin, Yang, Xie, Yang, Liu,
  Lin, Huang, and Zhou]{zhang2025qwenembedding}
Yanzhao Zhang, Mingxin Li, Dingkun Long, Xin Zhang, Huan Lin, Baosong Yang,
  Pengjun Xie, An~Yang, Dayiheng Liu, Junyang Lin, Fei Huang, and Jingren Zhou.
\newblock {Qwen3 Embedding: Advancing Text Embedding and Reranking Through
  Foundation Models}, 2025.
\newblock URL \url{https://arxiv.org/abs/2506.05176}.
\newblock arXiv:2506.05176.

\bibitem[Zhao et~al.(2025{\natexlab{a}})Zhao, Wu, Wu, Xu, Yue, Lin, Wang, Wu,
  Zheng, and Huang]{zhao2025absolutezero}
Andrew Zhao, Yiran Wu, Tong Wu, Quentin Xu, Yang Yue, Matthieu Lin, Shenzhi
  Wang, Qingyun Wu, Zilong Zheng, and Gao Huang.
\newblock {Absolute Zero: Reinforced Self-play Reasoning with Zero Data}.
\newblock In D.~Belgrave, C.~Zhang, H.~Lin, R.~Pascanu, P.~Koniusz,
  M.~Ghassemi, and N.~Chen, editors, \emph{Advances in Neural Information
  Processing Systems}, volume 38, Main Conference, pages 105816--105879. Curran
  Associates, Inc., 2025{\natexlab{a}}.
\newblock \doi{10.52202/085713-3534}.
\newblock URL
  \url{https://proceedings.neurips.cc/paper_files/paper/2025/file/9837dc00ff67d176373268ed48042d49-Paper-Conference.pdf}.

\bibitem[Zhao et~al.(2025{\natexlab{b}})Zhao, Wu, Guan, and
  Kong]{zhao2025promptcot}
Xueliang Zhao, Wei Wu, Jian Guan, and Lingpeng Kong.
\newblock {{P}rompt{C}o{T}: Synthesizing Olympiad-level Problems for
  Mathematical Reasoning in Large Language Models}.
\newblock In Wanxiang Che, Joyce Nabende, Ekaterina Shutova, and Mohammad~Taher
  Pilehvar, editors, \emph{Findings of the {A}ssociation for {C}omputational
  {L}inguistics: {ACL} 2025}, pages 18167--18188, Vienna, Austria, jul
  2025{\natexlab{b}}. Association for Computational Linguistics.
\newblock ISBN 979-8-89176-256-5.
\newblock \doi{10.18653/v1/2025.findings-acl.935}.
\newblock URL \url{https://aclanthology.org/2025.findings-acl.935/}.

\end{thebibliography}

\clearpage
\appendix

\raggedbottom
\setlength{\intextsep}{8pt plus 2pt minus 2pt}
\section{Implementation Details}
\label{app:implementation-details}

\paragraph{Software stack.}
{Both tasks use Qwen3-4B-Base, with \texttt{verl} for cold-SFT and GRPO-style reinforcement learning and vLLM for generation.}

\paragraph{Compute.}
{Experiments use eight H200 GPUs (140\,GiB each), 256 CPU cores, and 1.5\,TiB RAM. Estimated compute totals 3,400 GPU-hours. Integration training takes approximately 63, 110, and 116 hours across three iterations on eight GPUs; general-math training takes 26 hours for the setter (four GPUs) and 18 hours for the solver (eight GPUs).}

\begin{algorithm}[!htbp]
\begingroup
\small
\caption{{VHG training across iterations ($T=3$ for integration; $T=1$ for general math).}}
\label{alg:vhg}
\begin{algorithmic}[1]
\Require Seed bank $\mathcal{S}_1$; cold-SFT setter $\mathrm{Q}_0$; base solver $\mathrm{S}_0$; verifier $\mathrm{V}$; auxiliary data $\mathcal{D}_{\mathrm{aux}}$; solver sampling budget $K$; iterations $T$
\State $\mathcal{A}_0 \gets \emptyset$
\For{$t=1,\ldots,T$}
    \State Split $\mathcal{S}_t$ into training and validation sets
    \State Initialize setter $\mathrm{Q}_{\theta} \gets \mathrm{Q}_{t-1}$; keep solver $\mathrm{S}_{t-1}$ frozen
    \For{each setter update}
        \State Sample a batch of training seeds $s_i$ and pairs $(x_i,y_i^\star)\sim\mathrm{Q}_{\theta}(\cdot\mid s_i)$
        \State $v_i \gets \mathrm{V}(x_i,y_i^\star)$ \Comment{validity gate}
        \State For $v_i=1$, estimate $a_i=\mathrm{Acc}_{\mathrm{S}_{t-1}}(x_i,y_i^\star)$ from $K$ completions; otherwise $a_i\gets0$
        \State $r_i \gets v_i(1-a_i)$ \Comment{setter reward}
        \State Update $\theta$ by GRPO using the batch and rewards $\{r_i\}$
    \EndFor
    \State Select $\mathrm{Q}_t$ by validation reward
    \State Pool generations $\mathcal{G}_t$ from saved setter checkpoints on the complete seed bank $\mathcal{S}_t$
    \State $\mathcal{A}_t \gets \operatorname{Dedup}\!\left(\mathcal{A}_{t-1}\cup\{(x,y^\star)\in\mathcal{G}_t:\mathrm{V}(x,y^\star)=1\}\right)$
    \State Reassess archive $\mathcal{A}_t$ with $\mathrm{S}_{t-1}$ and retain eligible problems $\mathcal{E}_t$
    \State Form $\mathcal{D}_t\gets\operatorname{Dedup}(\mathcal{E}_t\cup\mathcal{D}_{\mathrm{aux}})$ and split into training and validation sets
    \State Train the solver from $\mathrm{S}_{t-1}$ on the training split of $\mathcal{D}_t$ with task-correctness reward $R_{\mathrm{S}}$
    \State Select $\mathrm{S}_t$ by validation reward
    \If{$t<T$}
        \State Prune unproductive seeds and expand $\mathcal{S}_{t+1}$ with problems from $\mathcal{A}_t$
    \EndIf
\EndFor
\State \Return $\{(\mathrm{Q}_t,\mathrm{S}_t)\}_{t=1}^{T},\;\mathcal{A}_T$
\end{algorithmic}
\endgroup
\end{algorithm}

{We randomly split cold-SFT and RL datasets into 80/20 training/validation sets in both tasks. Table~\ref{tab:implementation-optimization} summarizes the shared optimization settings. Cold-SFT runs for five epochs. Solver checkpoints are selected by validation reward.}

\begin{table}[H]

\caption{{Shared training settings. Length limits apply to complete SFT sequences and RL responses, respectively.}}
\label{tab:implementation-optimization}
\centering\small
\setlength{\tabcolsep}{4pt}
\begin{tabular}{l|rrrr}
\toprule
Stage & Learning rate & Prompt batch & Rollouts/prompt & Token limit \\
\midrule
Setter cold-SFT & $1{\times}10^{-5}$ & 32 & --- & 4,096 \\
Setter RL & $2{\times}10^{-6}$ & 128 & 8 & 8,192 \\
Solver RL & $2{\times}10^{-6}$ & 128 & 8 & 8,192 \\
\bottomrule
\end{tabular}
\end{table}

\paragraph{Verifier and filtering settings.}
{Local pass rate is estimated from ten solver completions per problem. For integration, the hard verifier checks the reference antiderivative against the generated integrand by differentiation.}

{{For general math, DeepSeek V3.2 supplies the soft-verifier reward during setter training.} We reject malformed outputs, near-copies of seeds, and degenerate answers. Solver-training candidates in the local pass-rate band $[0.1,0.9]$ are then verified by GPT-5.4. {See prompts in Appendix~\ref{app:soft-verifier-prompts}.}}

\paragraph{Evaluation settings.}
\begingroup
We report pass@1 and pass@8, with pass@1 as the primary metric. For $c$ correct responses among $n$ completions, the unbiased per-problem estimator~\citep{chen2021code} is
\begin{equation}
\widehat{\operatorname{pass@}k}=1-\frac{\binom{n-c}{k}}{\binom{n}{k}},
\label{eq:passk-extension}
\end{equation}
with value one when $n-c<k$. Scores average over problems, including aggregate integration scores.
\endgroup

{Integration benchmarks and OOD evaluations of integration-trained solvers use 64 completions per problem, temperature 0.7, and top-$p$ 0.95. Integration benchmarks allow 4,096 output tokens; OOD output limits are given in Appendix~\ref{app:ood}. General-math benchmark evaluations use 16 completions per problem, with 4,096 output tokens for MATH, GSM8K, Minerva, and Olympiad, and 8,192 for AMC and AIME. Model-scaling settings are given in Appendix~\ref{app:scaling-protocol}.}

{The Integral Stress Test contains human-authored problems from textbooks and online sources, normalized using OCR and LLM assistance. References are verified by differentiation, and overlaps with the integration seed set are removed. The benchmark is reserved for evaluation.}

\begingroup
\subsection{Additional integration iterations}
\label{app:multiround}

\paragraph{Initialization and model selection.}
Each additional iteration initializes the setter and solver from the preceding iteration and keeps the preceding solver frozen during setter training. Setter checkpoints are selected by validation reward, and each selected solver is used across all benchmarks. Each iteration uses one training run; intermediate benchmark evaluations were available during development.

\paragraph{Seed expansion.}
We retain seeds that previously yielded verified problems with solver pass rate below one, then deduplicate antiderivative--integrand pairs using exact expressions and numerical fingerprints. We embed these pairs with Qwen3-Embedding-8B~\citep{zhang2025qwenembedding} and select additions from the verified archive by farthest-first ranking relative to retained seeds and additions already selected. For $N$ retained seeds, we add $\lceil N/2\rceil$ candidates without a similarity cutoff. Only the antiderivative is supplied to the setter.

\paragraph{Generation and cumulative reuse.}
Each iteration pools generations across setter checkpoints and the complete seed bank. New verified problems are deduplicated into a cumulative archive and reassessed with the current frozen solver. Eligibility follows the original hardness check and local pass-rate band $[0.1,0.9]$. Previously used training problems remain eligible without reserved replay slots.

The eligible pool is combined with the same auxiliary dataset used in iteration one and deduplicated. The expanded seed bank and assembled solver dataset follow the shared training/validation split. The comparisons do not control for total training compute.

\endgroup

\section{Additional Experiment Results}
\label{app:additional-results}

\subsection{\texorpdfstring{{General-math benchmarks}}{General-math benchmarks}}
{Table~\ref{tab:math-results} reports the complete general-math results and data ablations.}

\begin{table}[H]
\caption{{General-math pass@1 results, including AIME 2026. Values are percentages; bold marks the best value in each column. R-Zero follows the authors' released implementation and evaluation protocol.}}
\label{tab:math-results}
\centering
\scriptsize
\setlength{\tabcolsep}{4pt}
\resizebox{\textwidth}{!}{%
\begin{tabular}{l|rrrrrrrrr}
\toprule
Method & MATH & GSM8K & AMC & Olympiad & Minerva & AIME24 & AIME25 & AIME26 & Overall p@1 \\
\midrule
Qwen3-4B-Base & 66.92 & 73.91 & 43.28 & 34.88 & 27.78 & 7.29 & 8.12 & 7.71 & 56.79 \\
Vanilla GRPO & 76.76 & 90.17 & 52.50 & 39.60 & 31.92 & \textbf{13.96} & 10.83 & 8.12 & 67.62 \\
\quad w. SFT data & 76.28 & 89.89 & 52.97 & 39.91 & 32.63 & 11.04 & \textbf{11.46} & 9.58 & 67.54 \\
R-Zero  &  &  &  & &  & &  &  &  \\
\quad Iter. 1 & 73.51 & 90.65 & 52.34 & 36.16 & 23.74 & 10.62 & 7.08 & 6.25 & 65.61 \\
\quad Iter. 2 & 73.59 & 91.37 & 49.84 & 35.41 & 28.52 & 10.00 & 7.71 & 7.50 & 66.20 \\
\quad Iter. 3 & 72.91 & \textbf{91.49} & 50.31 & 34.06 & 28.12 & 10.83 & 4.17 & 6.67 & 65.76 \\
\textbf{{\method} (Soft)} & \textbf{78.99} & 90.61 & \textbf{55.31} & \textbf{42.13} & \textbf{33.32} & 13.12 & \textbf{11.46} & \textbf{12.92} & \textbf{69.01} \\
\bottomrule
\end{tabular}}
\end{table}

\begingroup
\subsection{Out-of-domain transfer}
\label{app:ood}

\paragraph{Evaluation protocol.}
We evaluate HumanEval+~\citep{liu2023evalplus}, LiveCodeBench~\citep{jain2024livecodebench}, GPQA-Diamond~\citep{rein2024gpqa}, CRUXEval-O~\citep{gu2024cruxeval}, and MMLU-Pro~\citep{wang2024mmlupro}, using the sampling settings in Appendix~\ref{app:implementation-details}. Maximum output lengths are 4,096 tokens for LiveCodeBench and GPQA-Diamond, and 2,048 for the remaining benchmarks. Evaluation uses plain completions without tool use, execution feedback, or repair. Generated code is executed only by an isolated scorer. Malformed or unparseable answers remain in the denominator and count as incorrect.

Table~\ref{tab:ood-integration} reports pass@1 results for integration-trained solvers.

\begin{table}[H]

\caption{{OOD pass@1 results of integration-trained solvers (\%). Changes compare $S_3$ with Base in percentage points, with paired-bootstrap 95\% confidence intervals.}}
\label{tab:ood-integration}
\centering\small
\setlength{\tabcolsep}{4pt}
\begin{tabular}{l|rrrrrr}
\toprule
Benchmark & Base & \shortstack{R-Zero\\(best iter.)} & $S_1$ & $S_2$ & $S_3$ & Change [95\% CI] \\
\midrule
HumanEval+ & 62.35 & 64.49 & 67.58 & \textbf{68.18} & 68.16 & $+5.81\ [+3.78,+7.86]$ \\
LiveCodeBench & 18.37 & 18.83 & 19.47 & 19.63 & \textbf{19.79} & $+1.42\ [+1.19,+1.66]$ \\
GPQA-Diamond & 34.65 & 36.49 & 37.40 & 39.39 & \textbf{39.63} & $+4.98\ [+3.23,+6.78]$ \\
CRUXEval-O & 53.57 & 55.30 & 56.22 & 56.98 & \textbf{57.34} & $+3.77\ [+3.15,+4.39]$ \\
MMLU-Pro & 49.24 & 49.89 & 50.74 & 51.21 & \textbf{51.41} & $+2.17\ [+2.04,+2.30]$ \\
\bottomrule
\end{tabular}
\end{table}

\subsection{Uncertainty estimation and solver comparisons}
\label{app:uncertainty}
\label{app:rebuttal-analyses}

We use paired percentile-bootstrap 95\% confidence intervals, resampling problem identities jointly across compared models. For the first-iteration comparisons in Table~\ref{tab:original-uncertainty}, a second stage resamples each method's completion outcomes within each problem. The additional integration and OOD comparisons in Tables~\ref{tab:round-uncertainty} and~\ref{tab:ood-integration} resample problems only, stratifying by benchmark for integration. These problem-only intervals condition on the sampled completions. All intervals condition on the selected checkpoints, exclude training-run variability, and are unadjusted for multiple comparisons.

\begin{table}[H]

\caption{{First-iteration VHG pass@1 gains over baselines (percentage points), with paired two-stage bootstrap 95\% confidence intervals. Each domain weights its benchmark problems equally.}}
\label{tab:original-uncertainty}
\centering\small
\setlength{\tabcolsep}{4pt}
\begin{tabular}{l|rr}
\toprule
Method & Integration & General math \\
\midrule
Qwen3-4B-Base & $+19.99\ [+18.42,+21.55]$ & $+12.22\ [+11.41,+13.03]$ \\
Vanilla GRPO & $+4.29\ [+2.82,+5.78]$ & $+1.39\ [+0.86,+1.93]$ \\
\quad w. SFT data & $+5.85\ [+4.62,+7.09]$ & $+1.47\ [+0.93,+2.01]$ \\
R-Zero & & \\
\quad Iter. 1 & $+12.10\ [+10.52,+13.69]$ & $+3.40\ [+2.58,+4.24]$ \\
\quad Iter. 2 & $+10.86\ [+9.19,+12.53]$ & $+2.81\ [+1.96,+3.67]$ \\
\quad Iter. 3 & $+12.35\ [+10.61,+14.09]$ & $+3.25\ [+2.37,+4.14]$ \\
\bottomrule
\end{tabular}
\end{table}

\begin{table}[H]

\caption{{Overall integration changes (percentage points) with paired-bootstrap 95\% confidence intervals.}}
\label{tab:round-uncertainty}
\centering\small
\setlength{\tabcolsep}{4pt}
\begin{tabular}{l|rr}
\toprule
Transition & Pass@1 change [95\% CI] & Pass@8 change [95\% CI] \\
\midrule
Base $\rightarrow S_1$ & $+19.99\ [+18.51,+21.50]$ & $+9.74\ [+8.06,+11.49]$ \\
$S_1\rightarrow S_2$ & $+2.91\ [+1.73,+4.04]$ & $+1.67\ [+0.63,+2.70]$ \\
$S_2\rightarrow S_3$ & $+0.84\ [+0.06,+1.64]$ & $-0.01\ [-0.80,+0.78]$ \\
\bottomrule
\end{tabular}
\end{table}

The first-iteration VHG solver has a positive aggregate advantage over each comparator in both tasks.

{Each annual AIME test set contains only 30 problems, so we pool AIME 2024--2026 to reduce uncertainty from the small test sets. Across the 90 problems, {\method} achieves 12.50\% pass@1, exceeding R-Zero's best overall iteration by 4.10 percentage points (95\% confidence interval: [1.53, 6.94]).}

\subsection{Model-scaling evaluation protocol}
\label{app:scaling-protocol}

We evaluate the original hybrid Qwen3-4B, 8B, 14B, and 32B models using the same frozen weights in thinking and non-thinking modes. Each iteration supplies a fixed pool of 500 verified, deduplicated problems with zero correct completions among ten samples from the preceding solver. Pool identities are shared across all evaluators. Selection is non-random, is not balanced across setter checkpoints, and permits overlap between iterations.

Non-thinking inference uses temperature 0.7, top-$p$ 0.95, no top-$k$ truncation, and a 4,096-token output limit. Thinking uses temperature 0.6, top-$p$ 0.95, top-$k$ 20, and a 32,768-token limit including reasoning and the final answer. Each problem receives eight completions. The integration verifier checks extracted final answers, excluding the reasoning segment; malformed and truncated responses remain in the denominator.

Comparisons between thinking and non-thinking modes also reflect their different output budgets and sampling settings.

\endgroup

\subsection{\texorpdfstring{{Representative challenge examples}}{Representative challenge examples}}
{Tables~\ref{tab:challenge_examples_integration} and \ref{tab:challenge_examples_math} show generated challenges and their seeds; none of the sampled Qwen3-4B-Base attempts solved these examples.}

\begin{table}[H]
\caption{Indefinite-integral challenge examples generated by {\method}.}
\label{tab:challenge_examples_integration}
\centering
\scriptsize
\setlength{\tabcolsep}{4pt}
\begin{tabularx}{\textwidth}{p{0.32\textwidth}X}
\toprule
Original seed function & Generated problem \\
\midrule
$3e^{\sqrt[3]{x}}\!\left(x^{2/3}-2\sqrt[3]{x}+2\right)$
& Find an antiderivative of
\[
\begin{aligned}
&e^{\sqrt[3]{x}}x^{-2/3}\log x\,q(x)+\frac{3e^{\sqrt[3]{x}}}{x}q(x)\\
&+3e^{\sqrt[3]{x}}\log x\left(\frac{2}{3}x^{-1/3}-\frac{2}{3}x^{-2/3}\right),
\end{aligned}
\]
where $q(x)=x^{2/3}-2\sqrt[3]{x}+2$. \\
\midrule
$\begin{aligned}[t]
&\frac{3}{2}(x+1)^{2/3}-3\sqrt[3]{x+1}\\
&\quad+3\log(\sqrt[3]{x+1}+1)
\end{aligned}$
& Find an antiderivative of
\[
\begin{aligned}
&\frac{1}{3}(x+1)^{-2/3}r(x)\\
&+(x+1)^{1/3}\!\left((x+1)^{-1/3}-(x+1)^{-2/3}\right.\\
&\hspace{5em}\left.+\frac{(x+1)^{-2/3}}{\sqrt[3]{x+1}+1}\right),
\end{aligned}
\]
where $r(x)$ is the seed function. \\
\bottomrule
\end{tabularx}
\end{table}

\begin{table}[H]
\caption{General-math challenge examples generated by {\method}.}
\label{tab:challenge_examples_math}
\centering
\scriptsize
\setlength{\tabcolsep}{4pt}

\begin{tabularx}{\textwidth}{XX}
\toprule
Original seed & Generated problem \\
\midrule
In triangle $ABC$,
\[
\begin{aligned}
\tan\!\left(\frac{B-C}{2}\right)\tan\frac{A}{2}&=\frac{1}{29},\\
\tan\!\left(\frac{C-A}{2}\right)\tan\frac{B}{2}&=\frac{1}{27}.
\end{aligned}
\]
Find $\tan\!\left(\frac{A-B}{2}\right)\tan\frac{C}{2}$.
& In triangle $ABC$,
\[
\begin{aligned}
\tan\!\left(\frac{B-C}{2}\right)\tan\frac{A}{2}&=\frac{1}{29},\\
\tan\!\left(\frac{C-A}{2}\right)\tan\frac{B}{2}&=\frac{1}{27}.
\end{aligned}
\]
The side lengths $a=BC$ and $b=CA$ also satisfy $a/13=b/15$. Find $\tan\!\left(\frac{A-B}{2}\right)\tan\frac{C}{2}$. \\
\midrule
A blue plane, a red plane, and a white plane are waiting to take off from an airport with two runways. Planes take off one at a time and may use either runway. In how many ways can the three takeoffs be scheduled?
& A blue plane, a red plane, a yellow plane, and a white plane are waiting to take off from an airport with two runways. Planes take off one at a time and may use either runway. In how many ways can the four takeoffs be scheduled such that each runway has at least one plane taking off? \\
\bottomrule
\end{tabularx}
\end{table}

\section{Generation Analysis Details}
\label{app:generation-analysis-details}

\subsection{\texorpdfstring{{Learning dynamics and validity}}{Learning dynamics and validity}}

\begin{figure}[H]
\centering
\includegraphics[width=\textwidth]{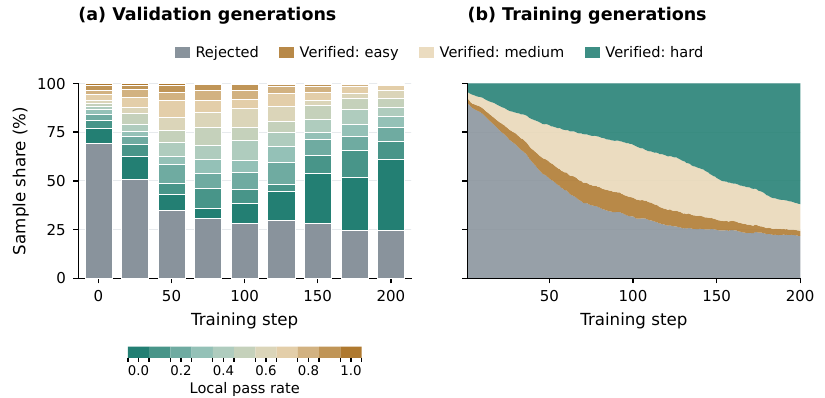}
\par\nointerlineskip
\caption{{Learning dynamics of integration generations. (a) Local pass-rate distributions at validation checkpoints; gray denotes rejected candidates. (b) Composition of training generations in rolling 10-step windows. Verified problems are grouped as hard ($p\leq0.3$), medium ($0.3<p<0.8$), or easy ($p\geq0.8$), where $p$ is local pass rate.}}
\label{fig:sym-hard-validation-pass-rate-distribution}
\label{fig:sym-hard-rollout-category-area}
\end{figure}

\begin{figure}[H]
\centering
\includegraphics[width=\textwidth,trim=0bp 3.25bp 0bp 0bp,clip]{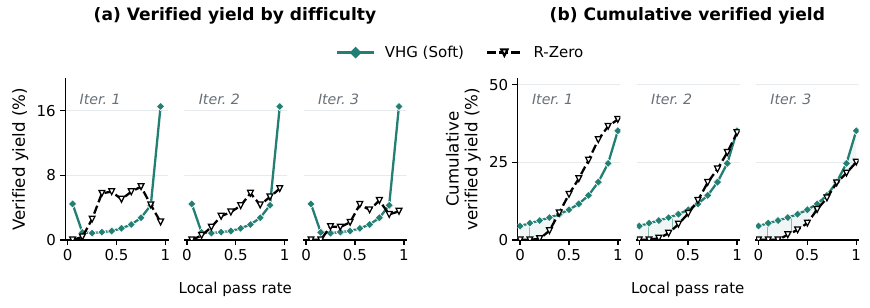}
\par\nointerlineskip
\caption{{Estimated verified yield for general-math generation under soft verification. (a) Verified yield within local pass-rate bins of width 0.1, with points at bin centers. (b) Cumulative verified yield as progressively easier problems are included, with markers at pass-rate increments of 0.1. Percentages are relative to each method's candidate pool and estimated by verifying samples from each bin. Each group compares the same VHG (Soft) pool against R-Zero iterations 1--3; iteration labels refer to R-Zero. Vertical hatching highlights where VHG is higher. Local pass rates retain each method's own difficulty measure.}}
\label{fig:general-math-judge-validity}
\end{figure}

\paragraph{Rollout-level novelty and reuse.}
{In integration training rollouts, 4.6\% of parsed generated references copy their seed, and 0.15\% of parsed integrands occur under multiple seeds. Validation generations have a 6.6\% seed-copy rate and no cross-seed integrand reuse.}

\begingroup
\subsection{Structural complexity, diversity, and seed dependence}
\label{app:diversity}

We analyze valid integration generations throughout all three setter iterations, generating ten problem-reference pairs per validation seed at each checkpoint. Seed banks are fixed within iterations and expanded between iterations. Repeated generations are retained.

Complexity is measured using SymPy expression trees: node count describes expression size and tree depth describes nesting. Figure~\ref{fig:integration-complexity-training} shows the complexity distributions during training.

\begin{figure}[H]
\centering
\includegraphics[width=\textwidth]{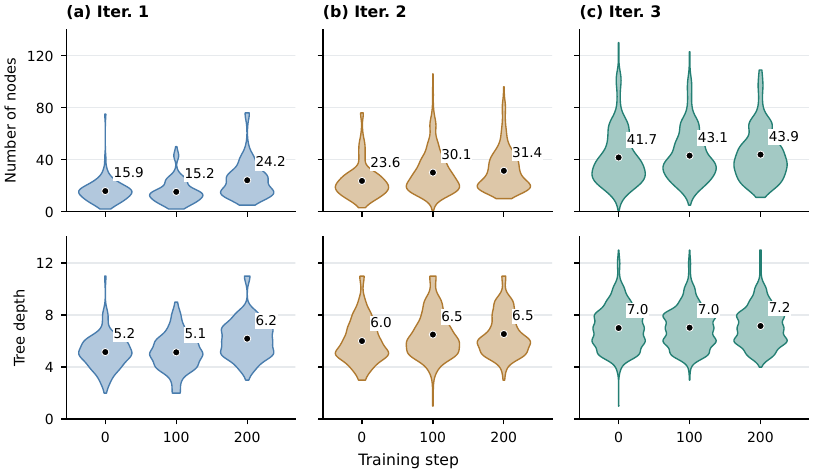}
\par\nointerlineskip
\caption{{Complexity distributions of valid integration generations at steps 0, 100, and 200 of each iteration. The top row shows expression size and the bottom row shows nesting depth, with common scales across iterations. Black points and labels mark means.}}
\label{fig:integration-complexity-training}
\end{figure}

\begin{samepage}
Grouping first-iteration generations by seed difficulty $d(s)=1-\widehat{\operatorname{pass@}1}(s)$ shows that hard seeds produce larger expression trees and lower solver pass rates (Figure~\ref{fig:integration-complexity-seeds}). This comparison is descriptive rather than a controlled intervention on seed difficulty.
\par\end{samepage}

\begin{figure}[H]
\centering
\includegraphics[width=\textwidth]{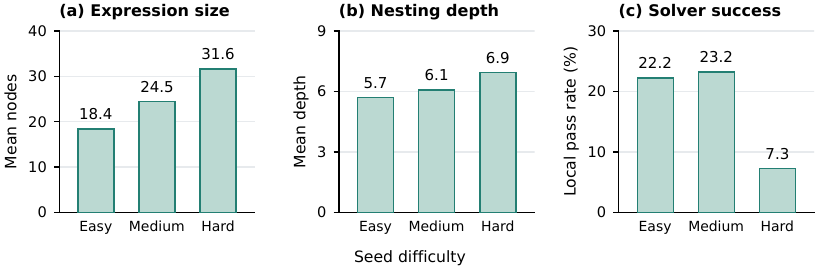}
\par\nointerlineskip
\caption{{Complexity and solver success at the final first-iteration checkpoint, grouped by seed difficulty: easy $[0,0.1]$, medium $[0.2,0.5]$, and hard $[0.6,1]$. Bars show means over valid generations from 14, 7, and 12 validation seeds, respectively.}}
\label{fig:integration-complexity-seeds}
\end{figure}

\paragraph{Integration method families.}
GPT-5.6 sol annotates integration methods and general-math diversity. Each valid first-iteration integration generation receives one primary method label. Figure~\ref{fig:integration-method-coverage} shows the resulting coverage before and after exact-expression deduplication.

\begin{figure}[H]
\centering
\includegraphics[width=\textwidth]{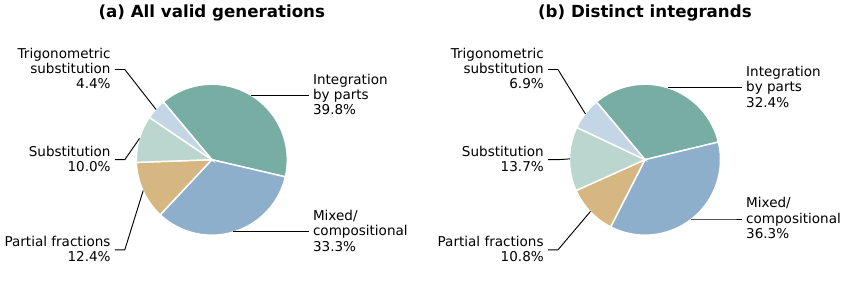}
\par\nointerlineskip
\caption{{Primary integration-method coverage in the first iteration, among (a) 249 valid generations and (b) 102 distinct integrands after exact-expression deduplication. Each problem contributes to one category.}}
\label{fig:integration-method-coverage}
\end{figure}

\begin{samepage}
\paragraph{General-math diversity.}
We uniformly sample 100 parseable problems accepted by the soft verifier at the final first-iteration general-math checkpoint, without replacement or prior deduplication. The same annotator labels their subjects, transformation types, and problem structures. The analysis identifies 99 distinct structures spanning six subjects and seven transformation types (Figure~\ref{fig:general-math-diversity-coverage}). These general-math annotations depend on the model and rubric and have not been independently validated by human annotators.
\par\end{samepage}

\endgroup

\begin{figure}[H]
\centering
\begin{minipage}{\linewidth}
{\small\textbf{(a) Setter}}\par\smallskip
\includegraphics[width=\linewidth]{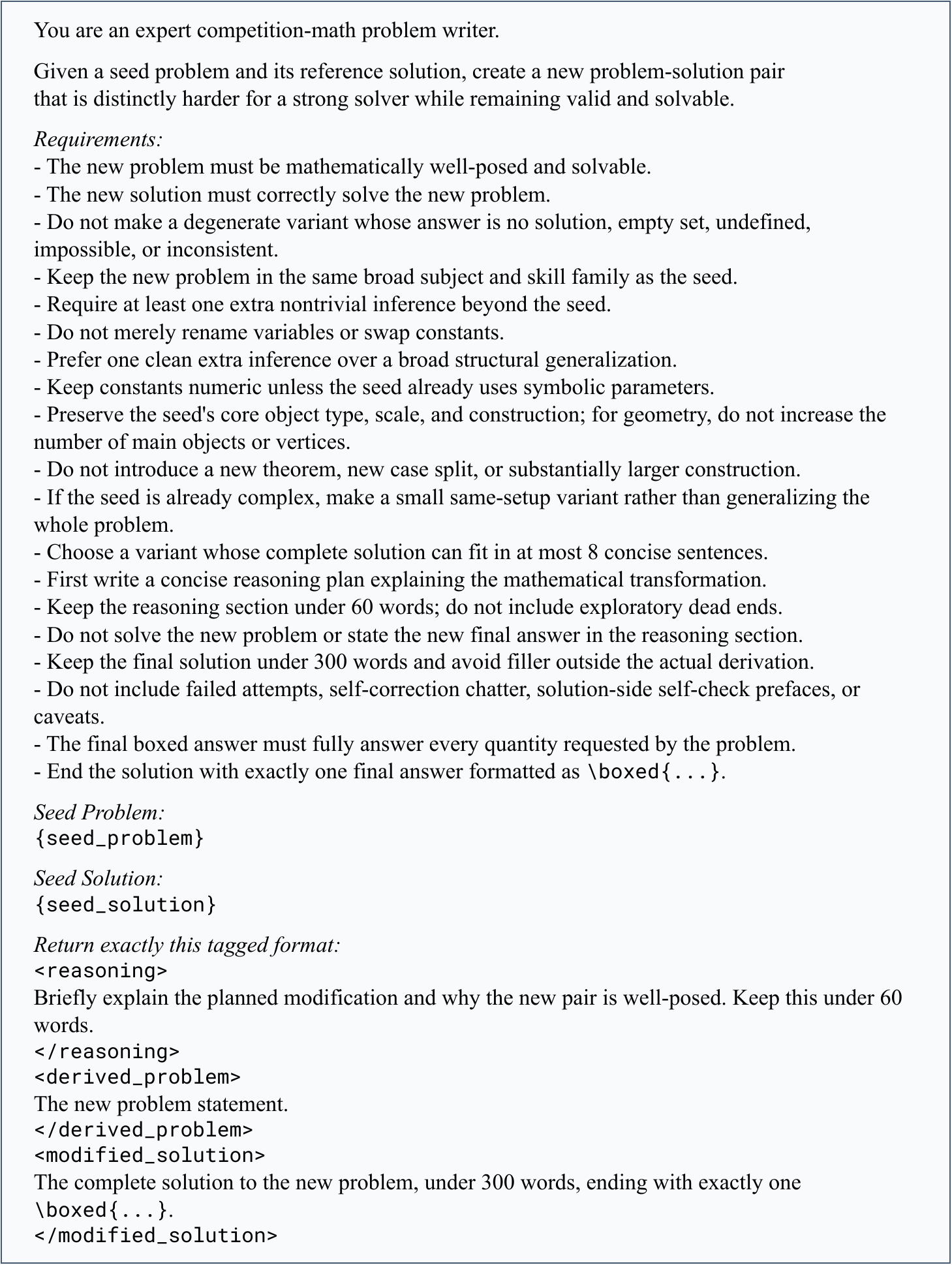}
\par\medskip
{\small\textbf{(b) Solver}}\par\smallskip
\includegraphics[width=\linewidth]{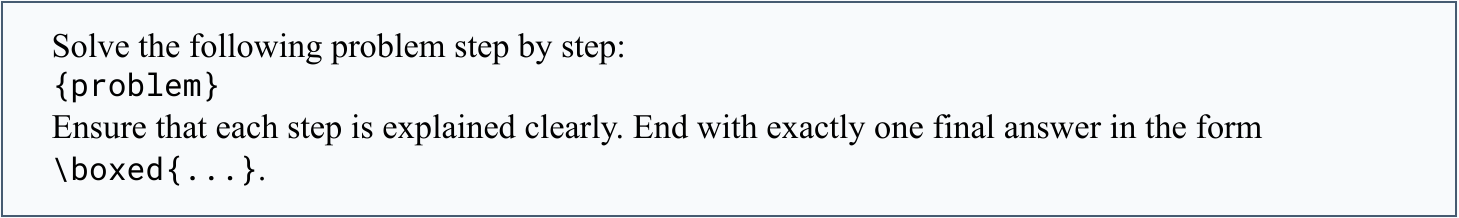}
\end{minipage}
\par\nointerlineskip
\caption{{Prompts for general math: (a) setter and (b) solver.}}
\label{fig:general-math-prompts}
\end{figure}

\begin{figure}[H]
\centering
\begin{minipage}{\linewidth}
{\small\textbf{(a) System message}}\par\smallskip
\includegraphics[width=\linewidth]{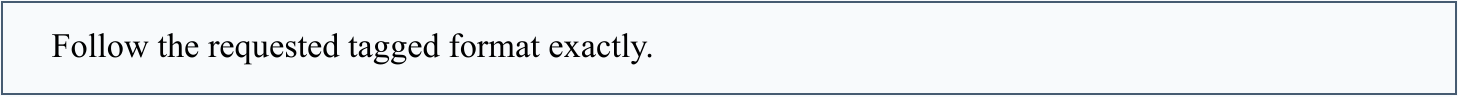}
\par\medskip
{\small\textbf{(b) User message}}\par\smallskip
\includegraphics[width=\linewidth]{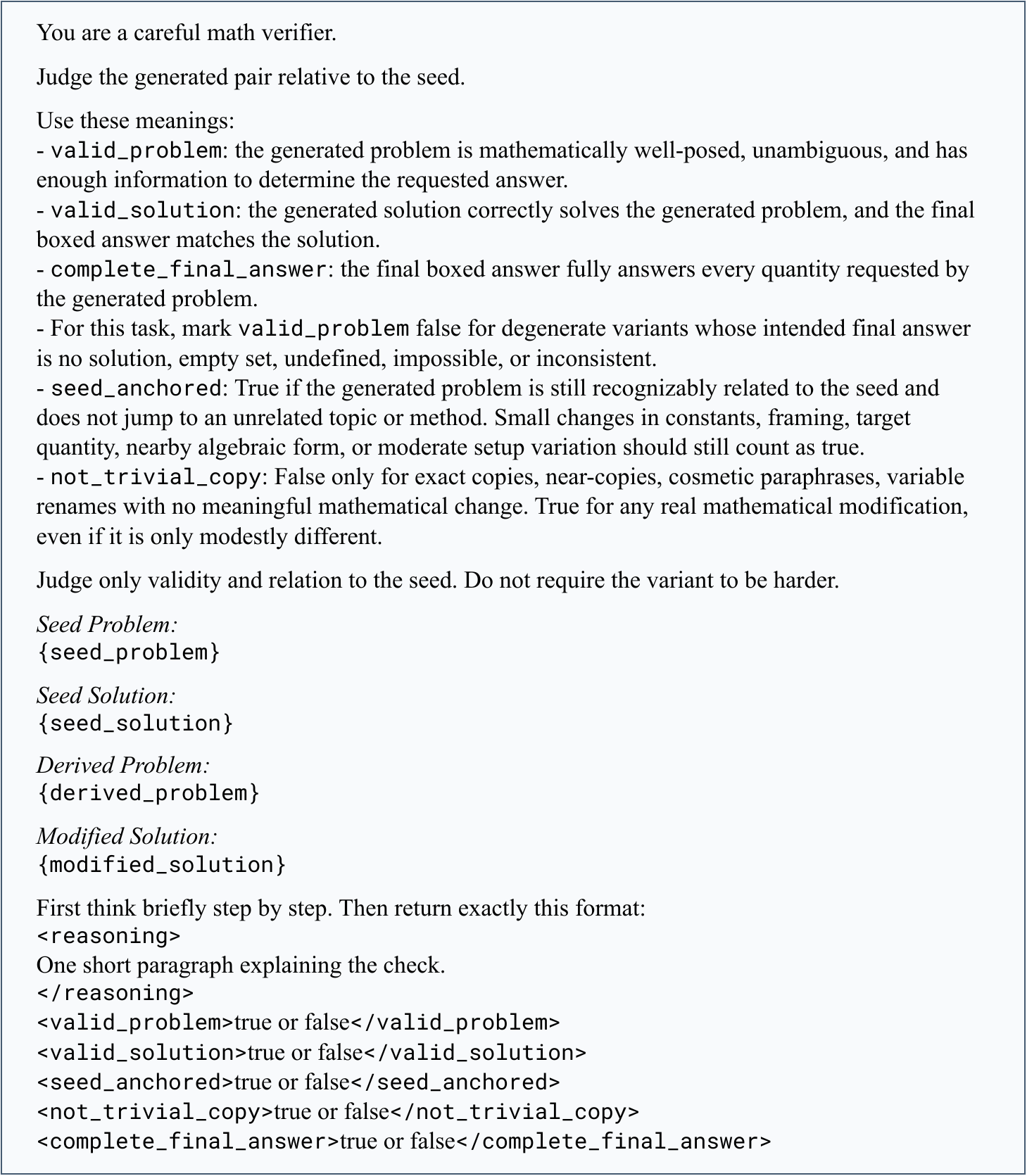}
\end{minipage}
\par\nointerlineskip
\caption{{Soft-verifier prompt: (a) system message and (b) user message.}}
\label{fig:soft-verifier-prompts}
\end{figure}

\begin{samepage}
\section{Prompts}
\label{app:prompts}
\label{app:setter-prompt}
\label{app:solver-prompt}
\label{app:soft-verifier-prompts}

{Figures~\ref{fig:integration-prompts}--\ref{fig:soft-verifier-prompts} present the setter, solver, and verifier prompts; Figures~\ref{fig:cold-sft-integration-1}--\ref{fig:cold-sft-math} present the teacher prompts used to collect cold-SFT demonstrations.}

\subsection{\texorpdfstring{{Setter, Solver, and Verifier Prompts}}{Setter, Solver, and Verifier Prompts}}

\begin{figure}[H]
\centering
\begin{minipage}{\linewidth}
{\small\textbf{(a) Setter}}\par\smallskip
\includegraphics[width=\linewidth]{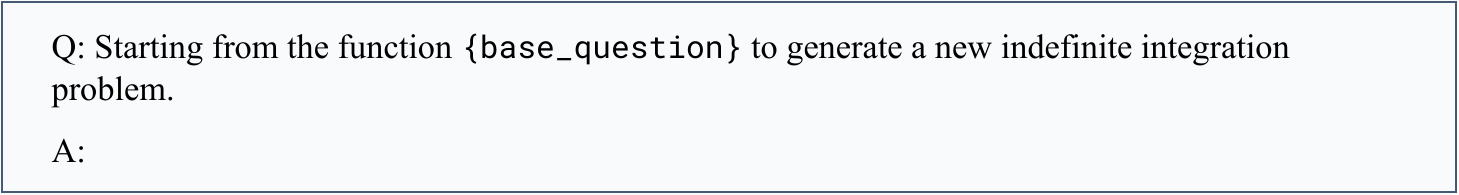}
\par\medskip
{\small\textbf{(b) Solver}}\par\smallskip
\includegraphics[width=\linewidth]{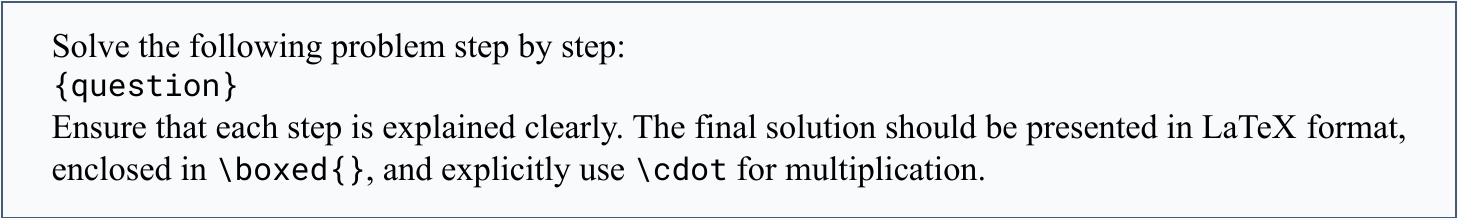}
\end{minipage}
\par\nointerlineskip
\caption{{Prompts for indefinite integration: (a) setter and (b) solver.}}
\label{fig:integration-prompts}
\end{figure}
\par\end{samepage}

\clearpage
\subsection{\texorpdfstring{{Cold-SFT Data Collection}}{Cold-SFT Data Collection}}
\begin{figure}[H]
\centering
\begin{minipage}{\linewidth}
{\small\textbf{(a) System message}}\par\smallskip
\includegraphics[width=\linewidth]{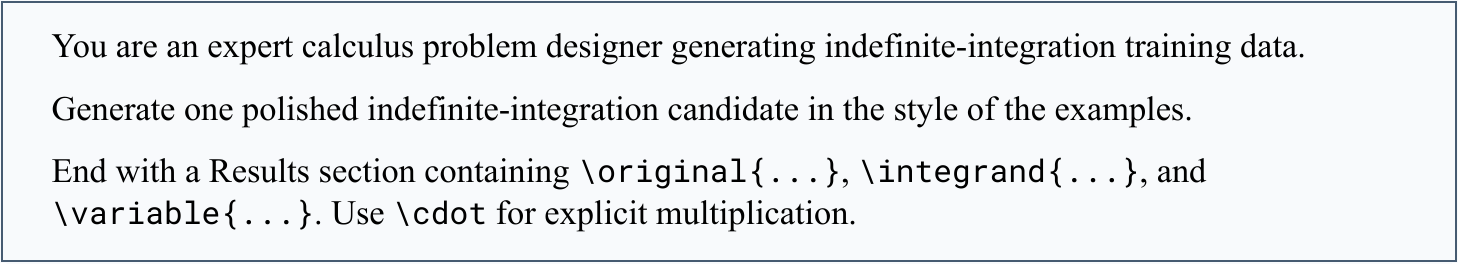}
\par\medskip
{\small\textbf{(b) User message, part 1 of 3}}\par\smallskip
\includegraphics[width=\linewidth]{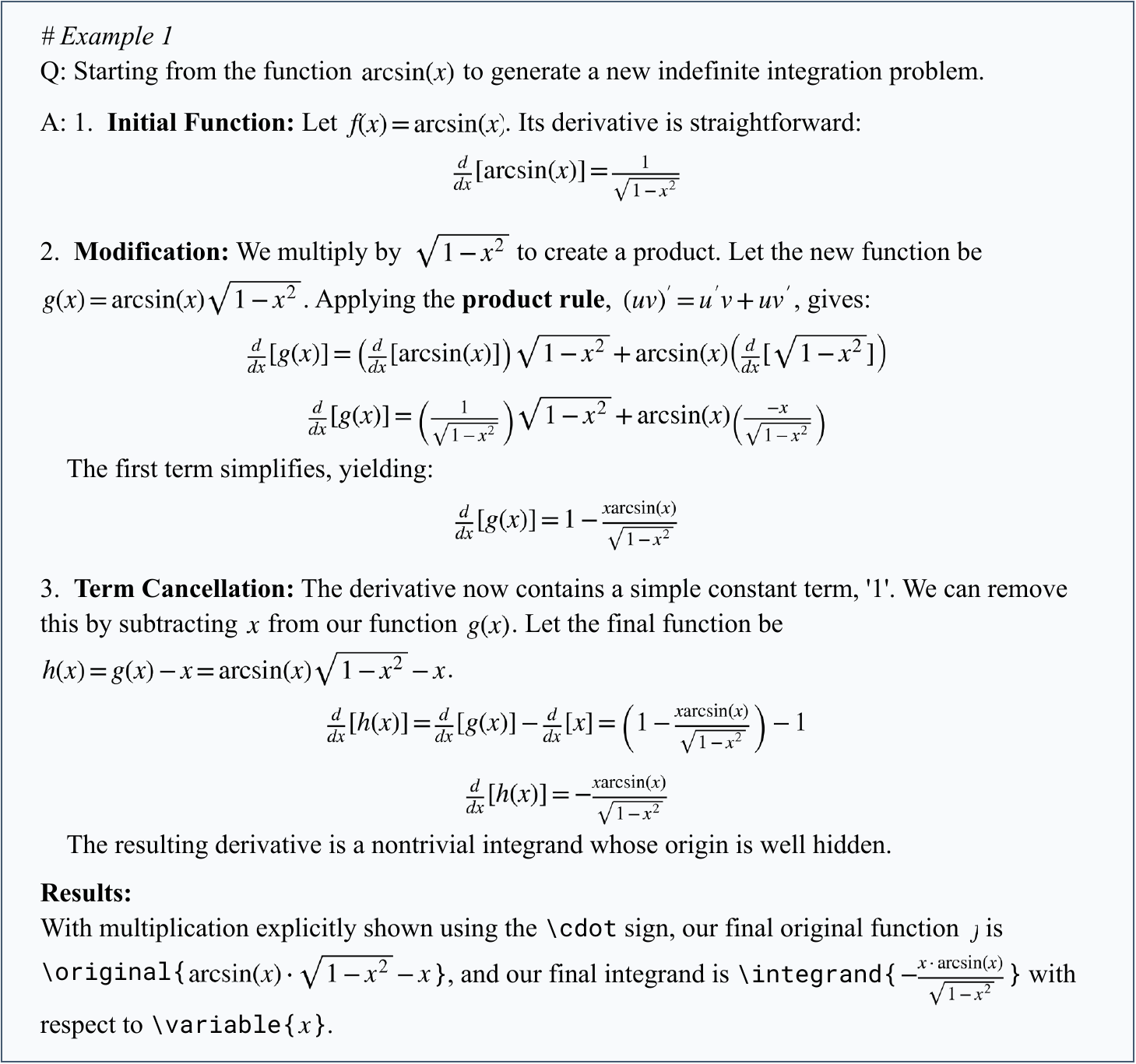}
\end{minipage}
\par\nointerlineskip
\caption{{Integration teacher prompt for cold-SFT: (a) system message and (b) user message, part 1 of 3. The user message continues in Figures~\ref{fig:cold-sft-integration-2}--\ref{fig:cold-sft-integration-3}.}}
\label{fig:cold-sft-integration-1}
\end{figure}

\begin{figure}[H]
\centering
\includegraphics[width=\linewidth]{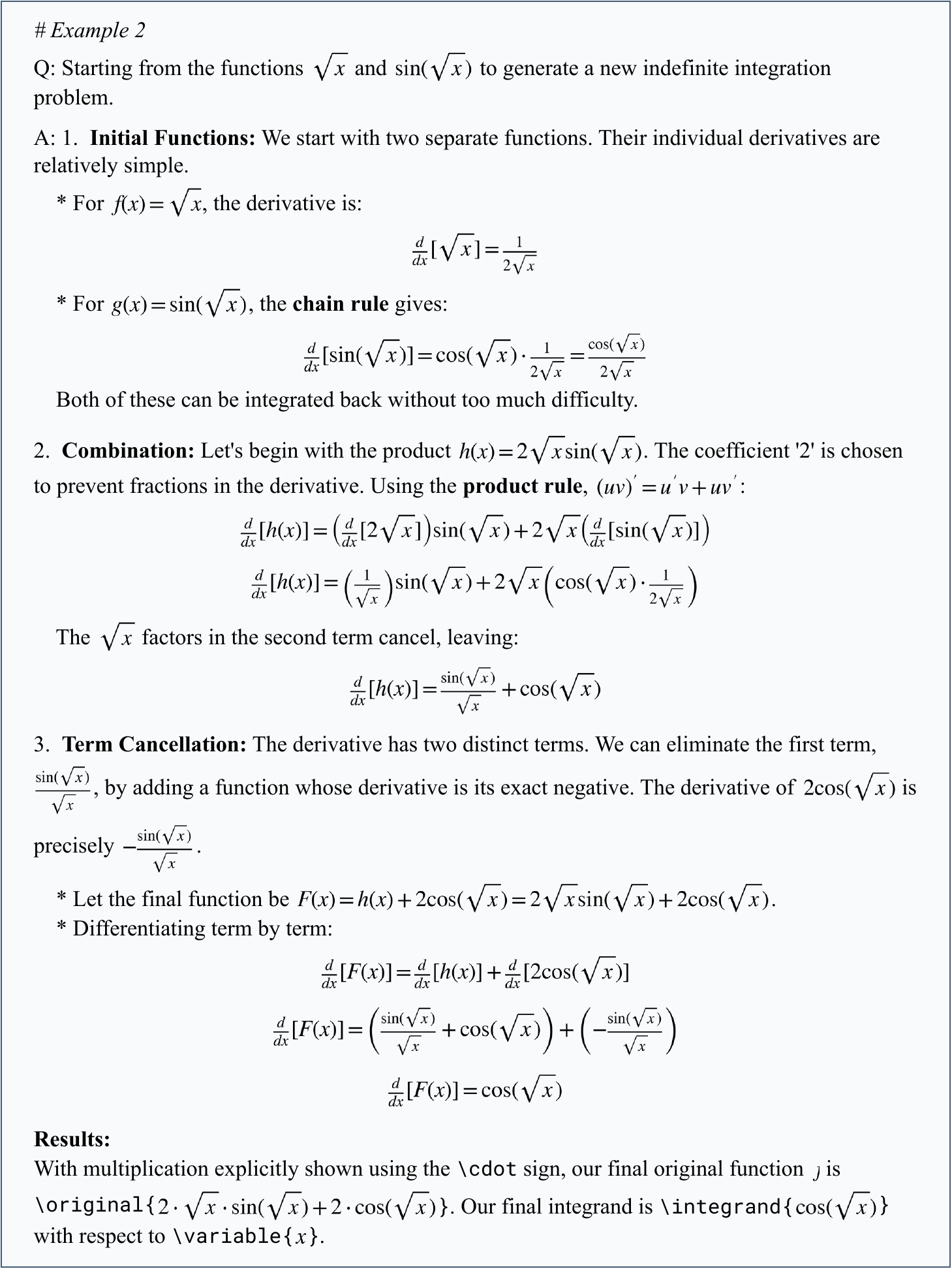}
\par\nointerlineskip
\caption{{Integration teacher prompt for cold-SFT, continued: user message, part 2 of 3.}}
\label{fig:cold-sft-integration-2}
\end{figure}

\begin{figure}[H]
\centering
\includegraphics[width=\linewidth]{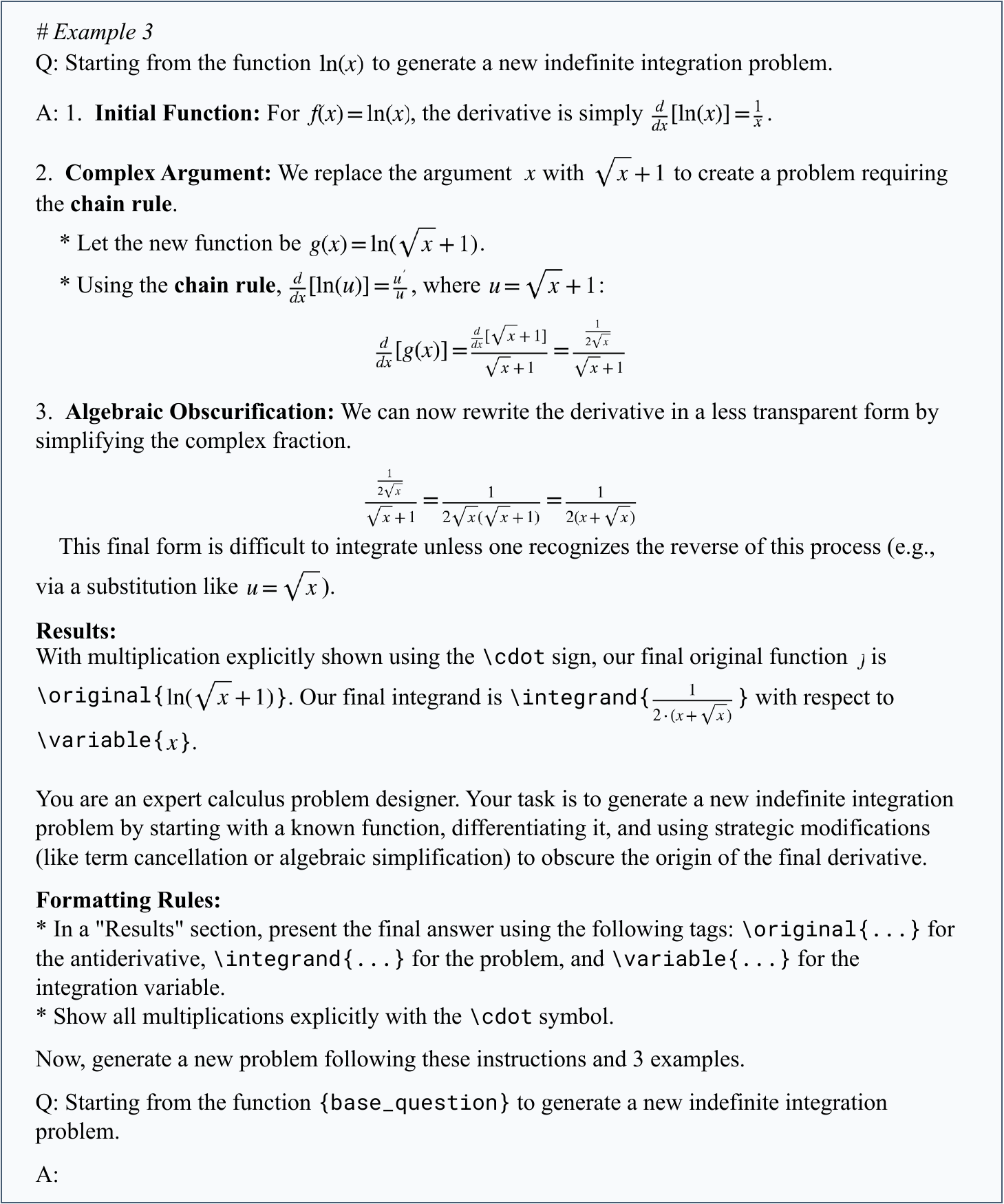}
\par\nointerlineskip
\caption{{Integration teacher prompt for cold-SFT, continued: user message, part 3 of 3.}}
\label{fig:cold-sft-integration-3}
\end{figure}

\begin{figure}[H]
\centering
\begin{minipage}{\linewidth}
{\small\textbf{(a) System message}}\par\smallskip
\includegraphics[width=\linewidth]{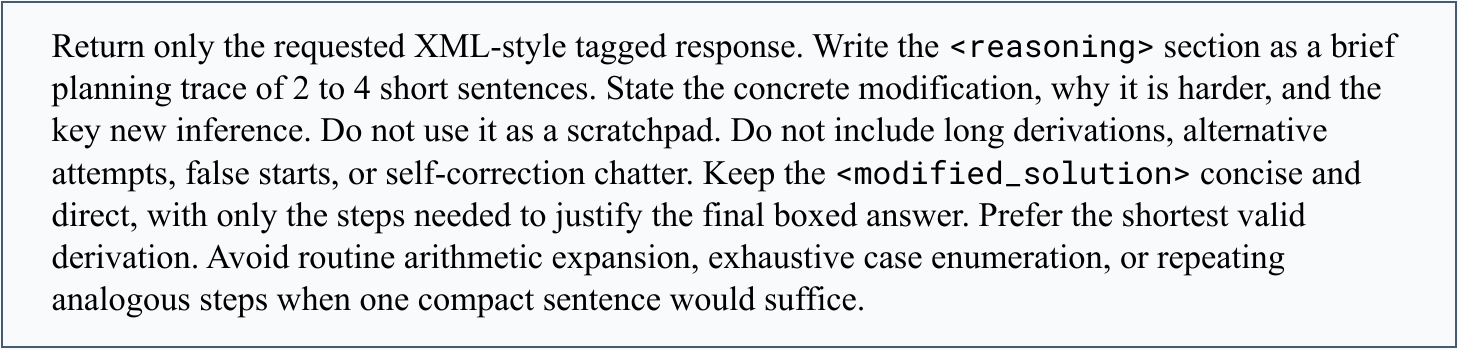}
\par\medskip
{\small\textbf{(b) User message}}\par\smallskip
\includegraphics[width=\linewidth]{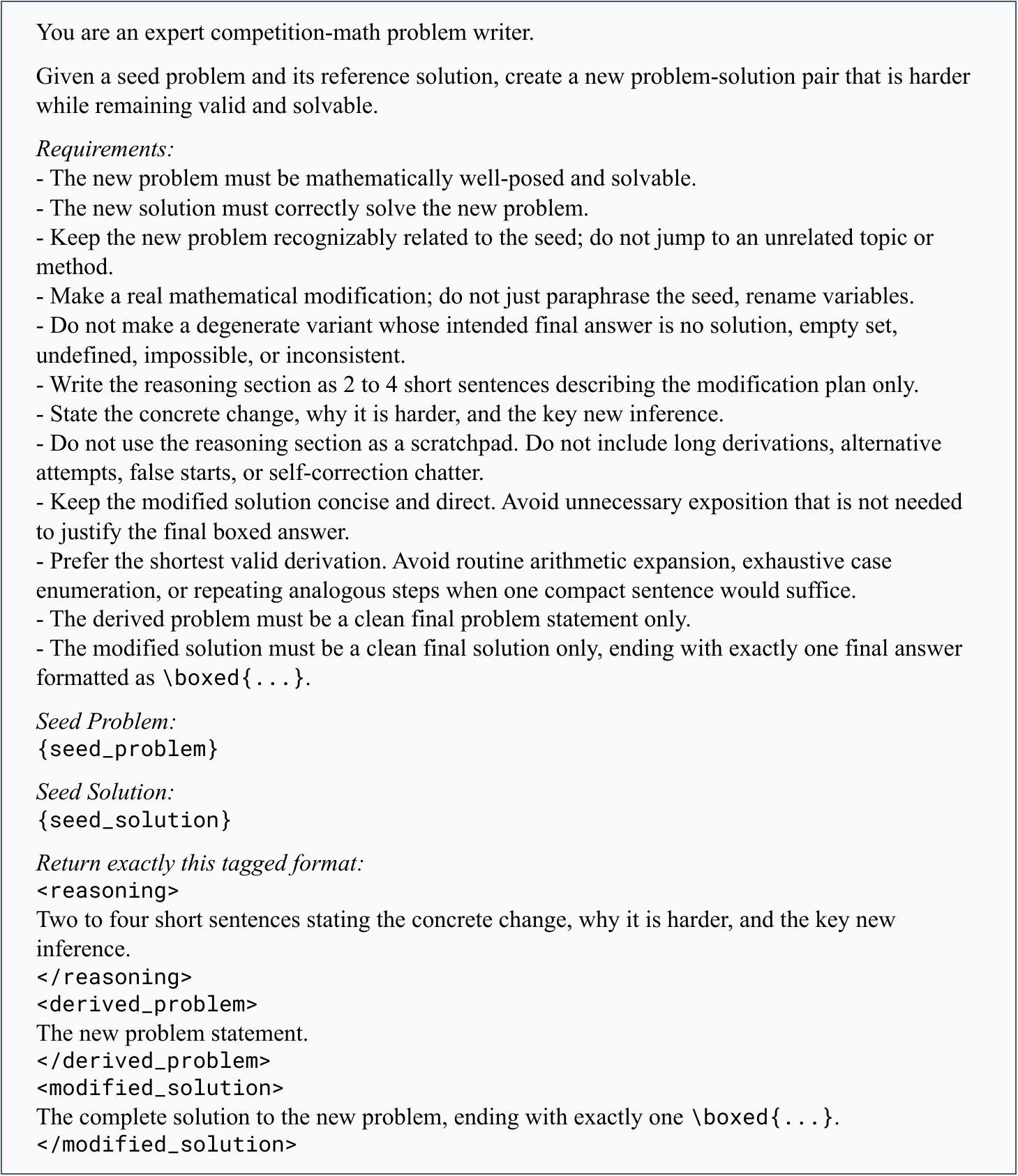}
\end{minipage}
\par\nointerlineskip
\caption{{General-math teacher prompt for cold-SFT: (a) system message and (b) user message.}}
\label{fig:cold-sft-math}
\end{figure}

{Cold-SFT uses the setter input templates in Figures~\ref{fig:integration-prompts}(a) and~\ref{fig:general-math-prompts}(a). Demonstrations for both tasks are collected from \texttt{deepseek-chat}. The integration teacher prompt spans Figures~\ref{fig:cold-sft-integration-1}--\ref{fig:cold-sft-integration-3}; the general-math teacher prompt appears in Figure~\ref{fig:cold-sft-math}.}

\par\noindent\begin{minipage}{\textwidth}
\setlength{\parskip}{5.5pt}
\section{Limitations and Broader Impact}
\label{sec:limitations}

{The guarantees of \method{} depend on its verifier. Indefinite integration provides a clean hard-verifier testbed within a narrow domain. General math is broader, but its LLM judge can accept subtle errors, underspecified problems, or reward-hacking artifacts. Our exact-validity claims therefore apply to the hard-verifier setting; the general-math results demonstrate a practical soft-verifier pipeline.}

{{We compare {\method} and R-Zero as complete systems, and fully matched configurations are not applicable.} Most experiments use one model family. Broader verifier backends, mathematical domains, and model families would strengthen the evidence for generality.}

{Verifier-backed generation can reduce noisy synthetic supervision and make hard-problem generation more transparent. It can also accelerate benchmark overfitting or produce misleading stress tests when verifier assumptions are undocumented. Explicit verifier documentation, separate hard- and soft-verifier claims, and independent validation are therefore important safeguards.}
\end{minipage}\par

\section{Asset Licenses}
\label{app:licenses}

\paragraph{Models and software.}
The \href{https://huggingface.co/Qwen}{Qwen3 and Qwen3-Embedding models}, \href{https://github.com/volcengine/verl/blob/main/LICENSE}{verl}, and \href{https://github.com/vllm-project/vllm/blob/main/LICENSE}{vLLM} use Apache 2.0; \href{https://github.com/sympy/sympy/blob/master/LICENSE}{SymPy} uses BSD-3-Clause. {\href{https://github.com/LiveCodeBench/LiveCodeBench/blob/main/LICENSE}{LiveCodeBench's evaluation code} uses MIT. The \href{https://github.com/Chengsong-Huang/R-Zero}{R-Zero repository} does not state an overall license; Apache-licensed inherited components do not specify terms for its own additions.}

\paragraph{Benchmark data.}
The \href{https://huggingface.co/datasets/hendrycks/competition_math}{MATH}, \href{https://huggingface.co/datasets/openai/gsm8k}{GSM8K}, \href{https://github.com/openai/human-eval/blob/master/LICENSE}{HumanEval}, \href{https://huggingface.co/datasets/cruxeval-org/cruxeval}{CRUXEval}, and \href{https://huggingface.co/datasets/TIGER-Lab/MMLU-Pro}{MMLU-Pro} releases list MIT licenses; \href{https://huggingface.co/datasets/evalplus/humanevalplus}{HumanEval+} uses Apache 2.0. \href{https://huggingface.co/datasets/Idavidrein/gpqa}{GPQA-Diamond} uses CC BY 4.0, with access terms prohibiting public disclosure of its examples. {The official \href{https://huggingface.co/datasets/Hothan/OlympiadBench}{OlympiadBench data release} and the \href{https://huggingface.co/datasets/project-numina/aimo-validation-aime}{upstream source of AIME 2024} list Apache 2.0. \href{https://huggingface.co/datasets/MathArena/aime_2026}{MathArena's AIME 2026 release} and the \href{https://ocw.mit.edu/pages/privacy-and-terms-of-use/}{MIT OpenCourseWare materials underlying Minerva} use CC BY-NC-SA 4.0. AMC and AIME problems originate from the MAA; the \href{https://huggingface.co/datasets/zwhe99/amc23}{AMC 2023} and \href{https://huggingface.co/datasets/yentinglin/aime_2025}{AIME 2025} mirrors used here do not declare licenses.}

{LiveCodeBench's authors describe academic evaluation under fair use in Appendix A.1 of their paper~\citep{jain2024livecodebench}; its dataset card leaves the Creative Commons variant unspecified. We use its problems only for evaluation.}

\paragraph{Integration problem sources.}
{\href{https://openreview.net/forum?id=8hRp0Ht108}{AntiderivBench} (Piotrowski and Yang, 2025) derives from \href{https://math.mit.edu/~yyao1/integrationbee.html}{MIT Integration Bee} and was obtained from its OpenReview supplement, which does not specify a dataset license. Other integration problems were collected from online sources without specified licenses, including digitized versions of \emph{Methods and Techniques of Integration} (Jin, Gu, and Mao, 2017) and \emph{Solutions to Demidovich's Problems in Mathematical Analysis}, Vol.\ 3 (Fei and Zhou, 4th ed., 2012).}

\end{document}